\documentclass[10pt,twocolumn,letterpaper]{article}

\usepackage{cvpr}      % To produce the REVIEW version
\usepackage{array}
\usepackage{tabularx}
\usepackage{ragged2e}
\usepackage[table]{xcolor}
\usepackage{booktabs}
\usepackage{amsmath}
\usepackage{amssymb}
\usepackage[normalem]{ulem}
\usepackage{algorithm}
\usepackage{algpseudocode}
\usepackage{adjustbox}
\usepackage{fontawesome5}
\definecolor{cvprblue}{rgb}{0.21,0.49,0.74}
\usepackage[pagebackref,breaklinks,colorlinks,allcolors=cvprblue]{hyperref}
\usepackage{twemojis}

\def\paperID{15694} % *** Enter the Paper ID here
\def\confName{CVPR}
\def\confYear{2026}

\title{Prototype-Guided Concept Erasure in Diffusion Models}

\author{Yuze Cai$^{1*}$ \quad Jiahao Lu$^{2*}$ \quad Hongxiang Shi$^{1}$ \quad Yichao Zhou$^{1}$ \quad Hong Lu$^{1\dagger}$\\
$^{1}$Fudan University\quad $^{2}$ National University of Singapore\\
{\tt\small
\begin{tabular}{c}
    yzcai24@m.fudan.edu.cn \quad jiahao.lu@u.nus.edu  \quad honglu@fudan.edu.cn
\end{tabular}
}
\\
{\small $^{*}$ Equal contribution \qquad $^\dagger$ Corresponding author}\\
{\small \faGithub \quad \url{https://github.com/Cocteau-23/Prototype-Guided-Concept-Erasure}}
}

\begin{document}
\twocolumn[{%
\renewcommand\twocolumn[1][]{#1}%
\maketitle
\vspace{-1.5em}
\includegraphics[width=\linewidth]{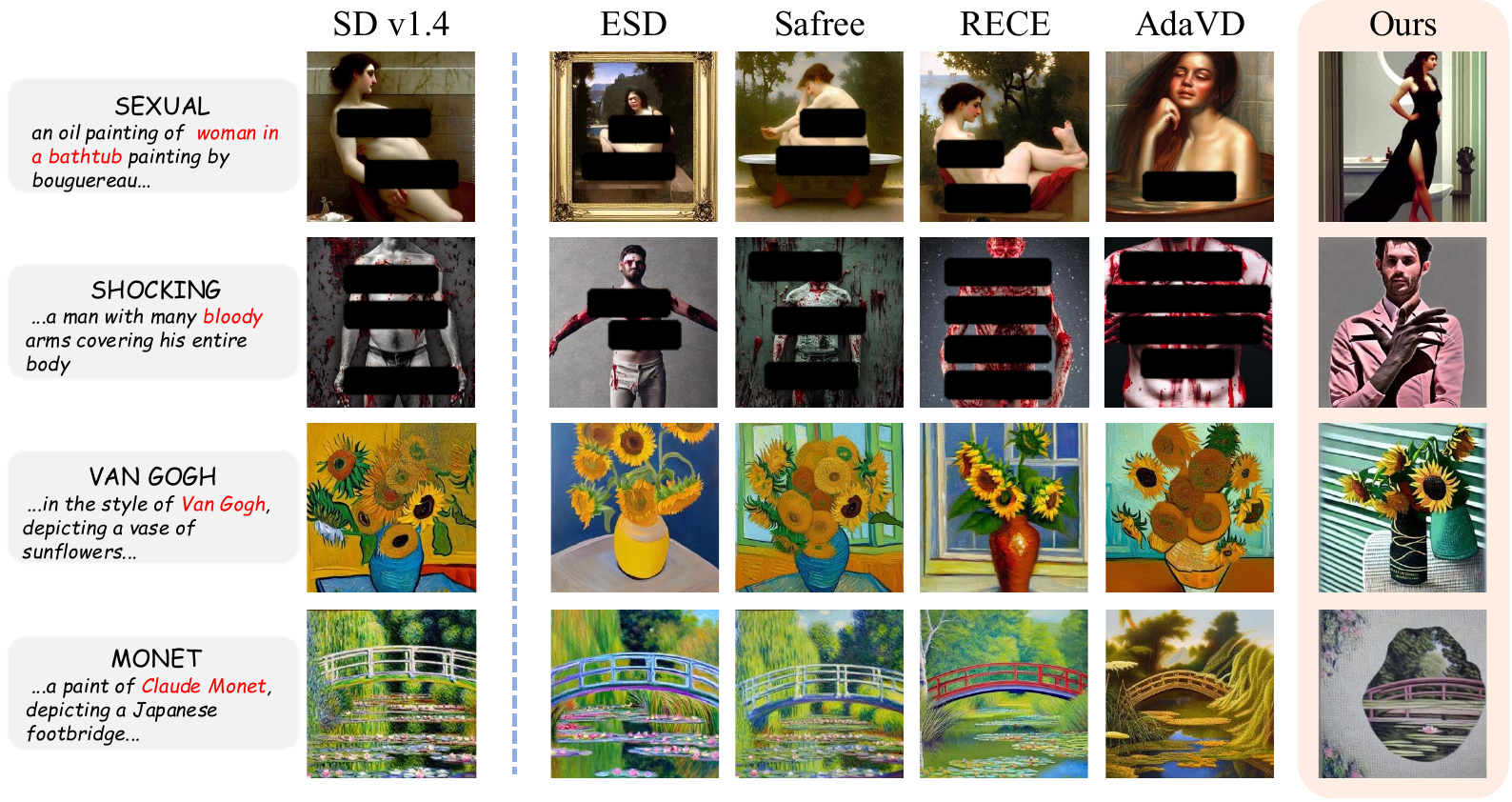}
\setlength{\abovecaptionskip}{-0.5em}
\setlength{\belowcaptionskip}{0pt}
\captionof{figure}{We present \textit{prototype-guided concept erasure}, a training-free method which models a target concept through a set of \textit{concept prototypes} that summarize its diverse semantic modes in the embedding space. 
These prototypes serve as negative guidance signals during inference and enable effective removal of both broad and narrow concepts while preserving the overall generation quality. 
Sensitive content is masked for display.}
\label{fig:teaser}
% \vspace{0.25em}
\noindent\textcolor{red}{\textbf{Content warning:} This paper contains content that may be inappropriate or offensive, such as violence, sexually explicit content, and negative stereotypes and actions.}
\vspace{1.5em}
}]

\begin{abstract}
    Concept erasure is extensively utilized in image generation to prevent text-to-image models from generating undesired content.  
    Existing methods can effectively erase narrow concepts that are specific and concrete, such as distinct intellectual properties (e.g. Pikachu) or recognizable characters (e.g. Elon Musk).
    However, their performance degrades on broad concepts such as ``sexual'' or ``violent'', whose wide scope and multi-faceted nature make them difficult to erase reliably.
    %Existing methodologies for model fine-tuning or parameter editing predominantly utilize text prompts as the condition for training or guidance. Although these approaches exhibit strong performance on specific, concrete concepts (e.g. distinct intellectual properties (IPs) or objects with clear visual forms), their efficacy is notably limited when addressing more abstract concepts, such as NSFW categories like 'sexual' or'nudity'. 
    To overcome this limitation, we exploit the model's intrinsic embedding geometry to identify latent embeddings that encode a given concept. 
    %Specifically, our methodology comprises three stages. First, we embed the concept-related images generated by the model into a text-image aligned multi-modal embedding space using an image encoder. Following, we perform clustering on these image embeddings to distill multiple cluster centroids that represent the concept.Second, we initialize a set of learnable concept prototypes, $P$, within the text embedding space, where each prototype $p_i \in P$ has the dimensions of $\mathbb{R}^{77 \times 768}$. These prototypes are then optimized to align with their corresponding cluster centroids, enabling $P$ to capture the core visual features associated with the target concept.Finally, during the inference stage, these learned concept prototypes serve as conditional inputs. Based on the given text prompt, the most relevant prototype is dynamically selected and utilized to apply negative guidance, thereby inhibiting the model from synthesizing inappropriate content.
    By clustering these embeddings, we derive a set of \textbf{concept prototypes} that summarize the model's internal representations of the concept, and employ them as negative conditioning signals during inference to achieve precise and reliable erasure.  
    %Therefore, we propose a method that employs image-based prompts to guide the model's generation process. We extract 'concept prototypes' from images generated by the model itself in response to the concept. These prototypes are subsequently used as negative conditioning signals at inference time, effectively preventing the model from producing content associated with the concept intended for erasure.
    Extensive experiments across multiple benchmarks show that our approach achieves substantially more reliable removal of broad concepts while preserving overall image quality, marking a step towards safer and more controllable image generation.

\end{abstract}

\section{Introduction}
\label{sec:Introduction}
Text-to-image (T2I) generation models ~\cite{dhariwal2021diffusion,ho2022classifier,podell2023sdxl,ho2020denoising,song2020denoising, wu2025qwen} have garnered significant attention for their capability to synthesize high-fidelity images from textual descriptions. 
However, as these models are typically trained on massive datasets containing uncurated web-scraped images, they inevitably risk learning undesirable concepts. 
Such concepts may include content associated with copyright infringement ~\cite{jiang2023ai, roose2022ai} or Not-Safe-For-Work (NSFW) content ~\cite{zhang2024generate}. 
This has, in turn, raised significant public concerns regarding the potential misuse of T2I models for generating inappropriate or harmful images.

\begin{figure*}[t!]
    \centering
    % \fbox{\rule{0pt}{1in} \rule{0.9\linewidth}{0pt}}
    \includegraphics[width=0.8\linewidth]{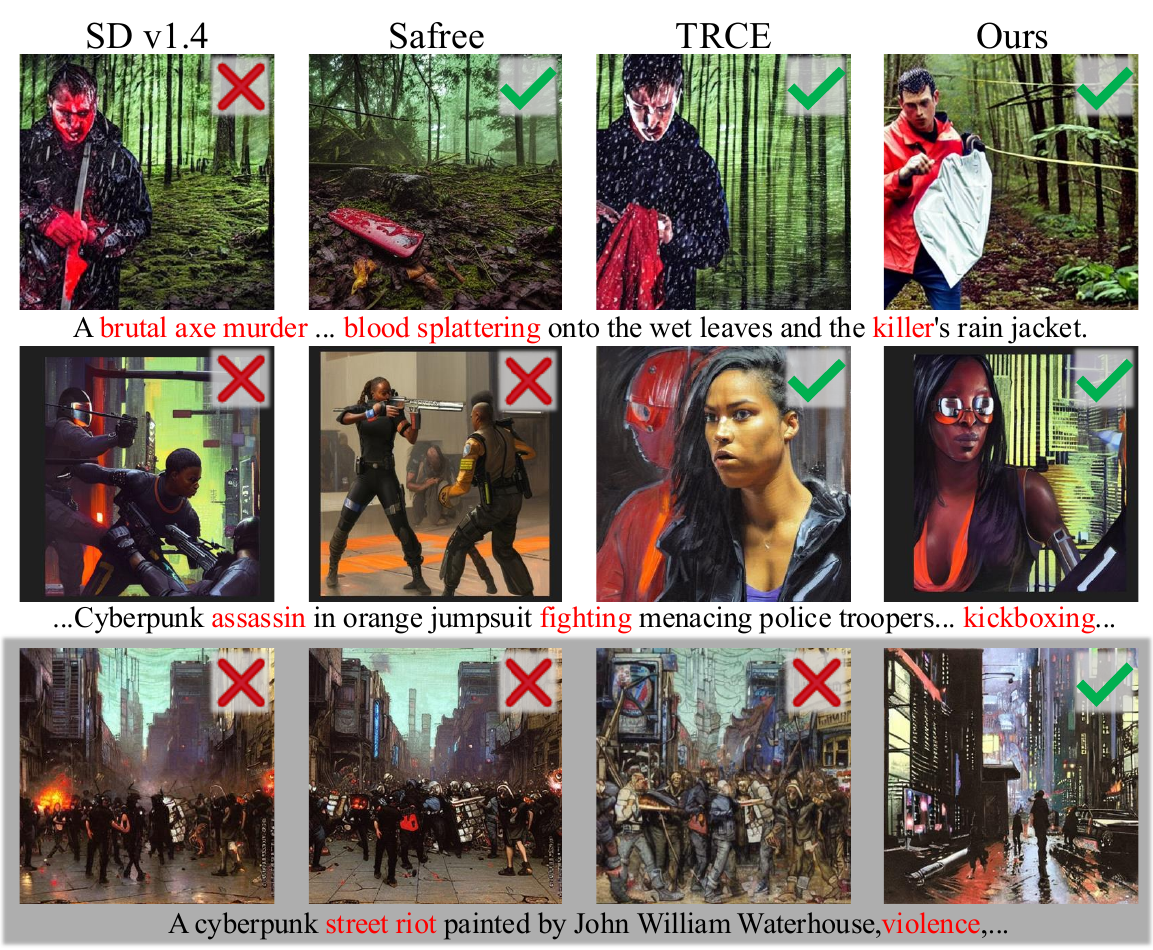}
    \caption{Broad concepts such as \textit{violence} encompass multiple semantic modes, including \textit{bloodshed} (first row), \textit{gunfights} (second row), and \textit{riots} (third row). Existing methods such as Safree~\cite{yoon2025safree} and TRCE~\cite{chen2025trce} erase only part of this spectrum, resulting in incomplete erasure. We highlight this unreliability as a core limitation of prior approaches and introduce our method, which achieves more comprehensive and reliable removal by explicitly modeling the full breadth of a target concept.}
    \label{fig:concept_breadth}
\end{figure*}

To address these risks, prior work has explored various strategies to enhance content safety of T2I models.
One line of work focuses on external safeguards such as filtering training corpora~\cite{rombach2022high} or attaching lightweight NSFW detectors to block unsafe outputs~\cite{rando2022red, schramowski2023safe}, which offer only coarse control over the generative models.
In contrast, \textbf{concept erasure} methods intervene directly in the model's output distribution, either by adjusting its parameters through targeted fine-tuning~\cite{kim2024race,li2024safegen,kumari2023ablating,schramowski2023safe,lu2024mace} or by steering the sampling process at inference time~\cite{gandikota2024unified,huang2024receler,gong2024reliable}, to suppress specific undesirable concepts.
Such concept erasure approaches typically provide stronger control with substantially lower computational cost than full model retraining, making them a promising basis for practical safety interventions.

Although existing concept erasure methods have shown promising results for removing narrow concepts such as specific intellectual properties (e.g., \textit{Pikachu}) or well-known individuals (e.g., \textit{Elon Musk}), they prove less capable when the target concept is broad.
Broad and open-ended concepts such as sexual or violent content can be realized through protean appearances: encompassing diverse visual forms across attire, pose, and environment,  while also exhibiting flexible textual expressions that vary with attributes and contexts.
The flexibility and high variety of such presentation forms makes broad concepts exceptionally difficult to erase completely, highlighting a key limitation of current erasure methods.
As illustrated in Fig.~\ref{fig:concept_breadth}, for broad concepts like \textit{violence}, existing techniques\cite{yoon2025safree, chen2025trce} primarily suppress the most overt instantiations (e.g., \textit{bloodshed}, first row) but fail to encompass other semantic modes such as \textit{gunfights} and \textit{riots} (second and third rows), leading to partial erasure.

A key reason is that prior works~\cite{gandikota2023erasing, kumari2023ablating, orgad2023editing, zhang2024forget, gandikota2024unified, lu2024mace, li2025one} implicitly assume uniformity between broad and narrow concepts, and model both with a single or unified signal.
While adequate for low-variance, narrowly defined concepts, this assumption breaks down for high-variance, multi-faceted concepts.

How to capture the many facets and high variability of broad concepts remains an underexplored problem.
Unlike narrow concepts, which are well represented by a specific name or concrete description, broad concepts such as ``hate'' or ``sexual'' encompass a wide range of textual expressions and visual forms, making it impractical to enumerate or collect all their possible manifestations.

Inspired by observations that generative models organize semantics into structured, low-dimensional neighborhoods rather than arbitrarily scatter~\cite{harkonen2020ganspace,shen2021closed,mokady2023null,gal2023image,tumanyan2023plug}, we posit that T2I models implicitly consolidate the representative modes of a concept within their embedding spaces.
These embedding spaces provide natural summaries of complex concepts, and benefit from robust cross-modal alignment between image and text.
Building on this intuition, we hypothesize that instances of a target concept reside in several compact regions of both the image and text embedding spaces.
We therefore summarize these regions using a set of representative anchors, which we term \textbf{concept prototypes}, each capturing a salient and expressive mode of the concept.
Ideally, these prototypes balance coverage and specificity, representing the full conceptual scope while providing precise semantic directions that enable training-free guidance in latent diffusion models.

To put this idea into practice, we introduce \textbf{prototype-guided concept erasure}, a training-free pipeline that captures concept diversity through a  set of learned \textit{concept prototypes}. We first construct image-space concept prototypes by comparing generations with and without the target concept, allowing us to capture the characteristic embedding shifts associated with each semantic mode. These image prototypes are then transferred into the text embedding space, producing textual prototypes that interface naturally with latent diffusion models.
During inference, we identify the prototype that aligns most closely with the user’s prompt and insert it as a negative conditioning signal within classifier-free guidance. This design allows the diffusion process to selectively down-weight the semantics encoded by the target concept, effectively suppressing it while maintaining fidelity, detail, and alignment for all unrelated aspects of the prompt.

Our main contributions can be summarized as follows:
\begin{enumerate}
    \item \textbf{We reveal a key weakness} of existing concept erasure methods: abstract concepts unfold multiple heterogeneous modes, which cannot be completely removed by treating them as a single direction.
    \item \textbf{We propose prototype-guided concept erasure}, a training-free  framework that captures the multimodal distribution of a concept through representative prototypes in both the image and text embedding spaces.
    \item \textbf{We demonstrate strong empirical performance} across multiple benchmarks, showing substantially improved effectiveness for removing broad concepts with minimal impact on overall generation quality.
\end{enumerate}

\section{Related Works}
\label{sec:Related_Work}

\textbf{Concept erasure} aims to thoroughly remove target concepts from pre-trained image generation models. 
Existing research works can be broadly categorized into two lines of work: training-based and training-free methods.
%To address this task, researchers have proposed numerous methods, which can be divided into two categories: training-based and training-free approaches. 

\noindent\textbf{Training-based methods} update the parameters of image generation models to erase undesired concepts.
\textit{Optimization-based approaches} removes the target concepts from model parameters by iteratively update from gradients.
%\emph{Optimization-based solutions} are straightforward to implement the parameter updates for erasing target concepts. 
Gandikota \textit{et al.}~\cite{gandikota2023erasing} achieve erasure through concept mapping, steering the noise estimation for the undesired concept towards a null or an anchor concept.
Kumari \textit{et al.} perform concept ablation~\cite{kumari2023ablating} by modifying the conditional distribution of generating target concept to a broader anchor concept.
FMN~\cite{zhang2024forget} fine-tunes cross-attention modules to eliminate certain concepts.
% \cite{gandikota2023erasing} achieve this via concept mapping, steering the noise estimation for the undesired concept towards a null or a desired target concept. While effective, these methods are often prone to inadvertently erasing semantically related concepts.
Li \textit{et al.}~\cite{li2025one} utilize text-image pairs as negative guidance to jointly suppress the generating probability of target concepts.
%More recent works, such as \cite{li2025one}, utilize text-image information to directly suppress the synthesis of the unwanted concept from a visual perspective.
To improve the robustness of concept erasure methods under adversarial setting, recent works such as Receler~\cite{huang2024receler}, AdvUnlearn~\cite{zhang2024defensive} and R.A.C.E~\cite{kim2024race} investigate erasure methods specifically against adversarial prompt attacks.

Apart from optimization-based training, some other methods derive \textit{closed-form solutions} to update model parameters without iterative training.
Notable closed-form methods include TIME~\cite{orgad2023editing} which modifies diffusion models' cross-attention layers to align outputs with desired attributes.
UCE~\cite{gandikota2024unified} and RECE~\cite{gong2024reliable} further refine the cross-attention weight update algorithms to achieve more precise removal.
MACE~\cite{lu2024mace} adopts a hybrid framework that computes a closed-form solution for multi-concept LoRA fusion, achieving scalable and precise concept removal.
%and uses a closed-form solution to solve the multi-concept lora fusion problem, achieving large-scale and precise concept erasure.
%Other approaches \cite{gandikota2024unified, gong2024reliable} formulate a closed-form solution to directly optimize the UNet cross-attention layers.  

\textbf{Training-free methods} avoid modifying model weights and thus enabling high adaptability and efficient deployment.
%Inference-time guidance-based methods do not alter the original model weights, 
They intervene in the diffusion denoising process to regulate the generative behaviors during inference time.
%instead performing dynamic intervention at each step of the UNet denoising process. 
%\cite{kumari2023ablating} guides the generation process by detecting the activation direction of the target concept within the UNet.
Safe Latent Diffusion~\cite{schramowski2023safe} applies redirection to the classifier-free guidance signal when current denoising trajectory is moving towards the undesired concept.
%\cite{schramowski2023safe} assesses whether the current generation trajectory is moving towards the undesired concept during the denoising process and applies guidance accordingly.
Free Safe Denoising~\cite{kim2025training} proposes an alternative formulation of denoising function itself to incorporate image-based unsafe priors to shift away from unsafe regions.
Wang \textit{et al.}~\cite{wang2025precise} performs value decomposition in the cross-attention layer. By projecting the value vector of the target concept onto the orthogonal complement space of the original prompt, it achieves fast and efficient erasing.
Other methods like Safree~\cite{yoon2025safree} steers text embedding away from toxic concept subspace to avoid generating harmful content.
%Other new methods, like \cite{yoon2024safree}, introduce perturbations to specific self-attention layers during unconditional noise prediction, thereby disrupting the model's capability to generate harmful concepts.

Our method falls under the category of training-free methods, relying solely on inference-time guidance to steer the denoising process away from undesirable concepts without modifying model weights.

\section{Prototype-Guided Concept Erasure}
\label{sec:methodology}

\begin{algorithm*}[t]
\caption{Prototype-Guided Concept Erasure}
\label{alg:main}
\begin{algorithmic}[1]

\State \textbf{Input:} Target concept $\kappa$, inference prompt $c$, parameters $(M,K, U,\tau,\alpha,\beta, T)$
\State \textbf{Output:} Generated image $x^{*}$ with concept $\kappa$ erased

\Statex \hrulefill
\State Collect concept-related prompts $\{c_i\}_{i=1}^N$
\State Construct concept-contrastive prompts $\{c_i^{-}\}_{i=1}^N$
\State Generate from LDM: $\mathcal{X}_i = \{x_{i,j}\}_{j=1}^M \sim c_i;\quad \mathcal{X}_i^{-} = \{x_{i,j}^{-}\}_{j=1}^M \sim c_i^{-}$
\State $z_{i,j} = \mathcal{E}_{\mathbf{I}}(x_{i,j}), \quad z_{i,j}^{-} = \mathcal{E}_{\mathbf{I}}(x_{i,j}^{-})$ \Comment{Encode images with CLIP}
\State $\mathcal{Z}_\mathrm{diff} = \{\,z_{i,j} - z_{i,k}^{-} \mid 1 \le i \le N,\; 1 \le j,k \le M\,\}$ \Comment{Embedding differences}
\State $\{p_{\mathbf{I}}^{(k)}\}_{k=1}^K = \mathsf{K}(\mathcal{Z}_\mathrm{diff})$ \Comment{Image concept prototypes}

\For{$k = 1,\dots,K$}
    \State $p_{\mathbf{T}}^{(k)} \sim \mathcal{N}(0,1)^{L\times d}$
    \For{$u = 1,\dots,U$}
        \State
        $p_{\mathbf{T}}^{(k)} \leftarrow
        p_{\mathbf{T}}^{(k)}
        +
        \eta\,
        \nabla_{p_{\mathbf{T}}^{(k)}}
        \frac{
            \langle p_{\mathbf{I}}^{(k)},\, \mathcal{E}(p_{\mathbf{T}}^{(k)}) \rangle
        }{
            \|p_{\mathbf{I}}^{(k)}\|\, \|\mathcal{E}(p_{\mathbf{T}}^{(k)})\|
        }$ \Comment{Optimize Eq.~\ref{eq:text_proto_optimization} for textual concept prototypes}
    \EndFor
\EndFor

\State $k^\ast = \arg\max_{k} \cos\!\big(\mathcal{E}(c),\, \mathcal{E}(p_{\mathbf{T}}^{(k)})\big)$
\State $z_T \sim \mathcal{N}(0,I)$
\For{$t = T$ \textbf{down to} $1$}    
    \State $\tilde{\epsilon}_\theta(z_t,c) \gets 
        \epsilon_\theta(z_t)
        + \alpha\big(\epsilon_\theta(z_t,c) - \epsilon_\theta(z_t)\big)
        - \beta\big(\epsilon_\theta(z_t,p_{\mathbf{T}}^{(k^\ast)}) - \epsilon_\theta(z_t)\big)$ 
    
    \State $z_{t-1} \gets 
        \frac{1}{\sqrt{\alpha_t}}
        \!\left(
            z_t
            -
            \frac{1 - \alpha_t}{\sqrt{1 - \bar{\alpha}_t}}
            \tilde{\epsilon}_\theta(z_t,c)
        \right)
        + \sigma_t\,\varepsilon,\quad \varepsilon\sim\mathcal{N}(0,I)$ \Comment{Reverse diffusion}
\EndFor

\State $x^{*} = \mathcal{D}_{\text{VAE}}(z_0)$
\State \Return $x^{*}$

\end{algorithmic}
\end{algorithm*}

In this section, we first review background on latent diffusion models, classifier-free guidance, and CLIP encoders (Sec.\ref{sec:preliminaries}). We then introduce our formulation of concept prototypes (Sec.\ref{sec:prototype_definition}), followed by how these prototypes are used as guidance signals to steer diffusion models away from the target concept during inference (Sec.~\ref{sec:method_erasure}).

\subsection{Preliminaries}
\label{sec:preliminaries}

Our method builds upon latent diffusion models~\cite{rombach2022high} guided by text embeddings and leverages the text–image alignment established by CLIP~\cite{radford2021learning}. We briefly review these foundations below.

\noindent\textbf{Latent Diffusion Models}. 
Diffusion models learn a target data distribution through iterative denoising~\cite{sohl2015deep, ho2020denoising}. 
Latent diffusion models (LDMs)~\cite{rombach2022high} improve efficiency by operating in the latent space.
Given an image $x$, its latent representation is encoded through a pretrained  variational autoencoder (VAE) encoder $z=\mathcal{E}_{\text{VAE}}(x)$, which is progressively noised into $z_t$ across timesteps $t$. 
An LDM parameterized by $\theta$ predicts the added noise $\epsilon_\theta(z_t, c)$ conditioned on an optional text prompt $c$. 
The training objective of an LDM is defined as:
\begin{equation}
    \mathcal{L}_{\text{LDM}} = 
    \mathbb{E}_{z_t, c, \epsilon \sim \mathcal{N}(0,1)} 
    \big[\|\epsilon - \epsilon_\theta(z_t, c)\|^2\big].
\end{equation}
During inference, $z_T \sim \mathcal{N}(0, I)$ is iteratively denoised to obtain $z_0$, which is eventually decoded back to image space as $x_0 = \mathcal{D}_{\text{VAE}}(z_0)$ through the image decoder of LDM.

\vspace{0.5em}
\noindent\textbf{Classifier-free Guidance} (CFG)~\cite{dhariwal2021diffusion} enables controllable generation without external classifiers. 
By contrasting the denoising predictions between the conditional and unconditional samplings, the adjusted score $\tilde{\epsilon}_\theta$ is guided with scale $\alpha$ towards the semantics specified by the text prompt $c$:
\begin{equation}
    \tilde{\epsilon}_\theta(z_t, c)
    = \epsilon_\theta(z_t)
    + \alpha\big(\epsilon_\theta(z_t, c) - \epsilon_\theta(z_t)\big),
\end{equation}

\noindent\textbf{CLIP Encoders}. 
LDMs commonly adopt its pretrained text encoder $\mathcal{E}_{\mathbf{T}}(\cdot)$ from CLIP~\cite{radford2021learning}, which provides a shared embedding space jointly learned with an image encoder $\mathcal{E}_{\mathbf{I}}(\cdot)$. CLIP’s large-scale contrastive pretraining yields robust cross-modal alignment, making its encoders well-suited for conditioning LDMs~\cite{ramesh2022hierarchical, saharia2022photorealistic} and for measuring image-text correspondence~\cite{hessel2021clipscore, hu2023tifa}. In practice, the cosine similarity between CLIP image and text embeddings is widely used as a standard metric of semantic alignment.

% \subsection{Heterogeneity of Concepts in Latent Space}
% \label{sec:concept_space_analysis}

% \TODO{Evidence for showing that single, unique directions cannot reliably present the broad concepts}.

% \TODO{Illustrating broad / narrow concept differences in the embedding space}

\subsection{Concept Prototype}
\label{sec:prototype_definition}

% As demonstrated in Sec.~\ref{sec:concept_space_analysis}, broad concepts exhibit heterogeneous structure that cannot be represented by a single direction in embedding space.  
Broad concepts exhibit high-variety and multi-faceted structure that cannot be represented by a single direction in the embedding space.
We thus capture their diversity through a set of \textit{concept prototypes} drawn from both the image and text embedding spaces.

Specifically, for a given target concept $\kappa$, we begin by collecting a set of text prompts containing related semantics of size $N$.
Since these prompts may include information irrelevant or orthogonal to $\kappa$, we construct a \textit{concept-contrastive prompt set} in which each prompt retains all contextual content but omits $\kappa$. 
For example, removing concept ``sexual'' from the following prompt changes its expression from red-highlighted to blue-highlighted text, while keeping other descriptive content unchanged:

\begin{quote}
\textit{documentary style an adult couple in a dressing room, \textcolor{red}{\sout{artistic nude silhouette}} \textcolor{blue}{casual clothing}, overcast soft light, over-the-shoulder shot.}
\end{quote}

For each prompt $c_i$ and its  concept-contrastive counterpart $c_i^{-}$, we  generate $M$ images with and without the target concept $\kappa$ respectively. 
Each image (either $x_{i, j}$ from $c_i$ or $x_{i, k}^{-}$  from $c_i^{-}$) is encoded by the CLIP image encoder, and we compute all pairwise differences between the two embedding sets:
\begin{equation}
    \mathcal{Z}_\mathrm{diff} =
\big\{\, z_{i,j} - z^{-}_{i,k}
\;\big|\;
1 \le i \le N,\; 1 \le j,k \le M
\big\}.
\end{equation}

To obtain representative semantic directions, we apply a clustering operator $\mathsf{K}(\cdot)$ to $\mathcal{Z}^\text{diff}$ to produce a set of \textit{image concept prototypes}:
\begin{equation}
    \{p_{\mathbf{I}}^{(1)}, \dots, p_{\mathbf{I}}^{(K)}\}
=
\mathsf{K}(\mathcal{Z}_\mathrm{diff}),
\end{equation}
where each prototype $p_{\mathbf{I}}^{(k)} \in \mathbb{R}^{d}$ is a cluster centroid capturing one expressive mode of the concept in the image embedding space of dimension $d$.

To enable direct control over LDMs, which are conditioned on text prompts, we have to transfer the image-space prototypes into text domain.
We construct a set of \textit{textual concept prototypes}, where each prototype is a learnable soft prompt $p_{\mathbf{T}}^{(k)} \in \mathbb{R}^{L \times d}$ consisting of $L$ learnable token embeddings, equivalently serving as a length-$L$ prompt for conditioning the LDM.
To perform this cross-modal transfer, we take advantage of  the pretrained CLIP text encoder, which offers strong text–image alignment and is already used as the conditioning module in standard LDMs.
To pair each textual prototype $p_{\mathbf{T}}^{(k)}$ with its counterpart image prototype $p_{\mathbf{I}}^{(k)}$, we consider the following optimization problem by maximizing their cosine similarity in the aligned embedding space:
\begin{equation}
\label{eq:text_proto_optimization}
    \max_{p_{\mathbf{T}}^{(k)}} 
    \;
    \frac{
        \langle p_{\mathbf{I}}^{(k)},\,
        \mathcal{E}(p_{\mathbf{T}}^{(k)}) \rangle
    }{
        \|p_{\mathbf{I}}^{(k)}\|\;
        \|\mathcal{E}(p_{\mathbf{T}}^{(k)})\|
    },
\end{equation}
where $\mathcal{E}(\cdot)$ is a differentiable mapping that projects a textual embedding into the joint CLIP embedding space. 
Specifically, $\mathcal{E}(p_{\mathbf{T}}^{(k)})$ is obtained by feeding $p_{\mathbf{T}}^{(k)}$ into the CLIP text encoder and extracting the End-of-Text (EoT) token embedding, which CLIP uses as a global summary of the entire sequence~\cite{chen2024training, li2024get, chen2025trce}.
Through the encoder’s attention layers, information from all token positions is naturally aggregated into this final embedding, allowing gradients to flow back through the frozen encoder and update only the learnable prompt parameters in $p_{\mathbf{T}}^{(k)}$.
After convergence, the optimized $p_{\mathbf{T}}^{(k)}$ serves as a text-space concept prototype representing the semantic mode associated with concept $\kappa$.

\subsection{Concept Erasure with Prototype Guidance}
\label{sec:method_erasure}
During inference, given a textual prompt $c$ that may contain the undesirable concept $\kappa$, we identify the prototype most relevant to the prompt.
Specifically, we compute the cosine similarity between the embedding of $c$ and each textual prototype, and select the top-1 prototype whose similarity exceeds a threshold $\tau$:

\begin{equation}
\begin{split}
    k^\ast
    &=
    \arg\max_{k \in \{1,\ldots,K\}}
    \;
    \cos\!\big(
        \mathcal{E}(c),\,
        \mathcal{E}(p_{\mathbf{T}}^{(k)})
    \big)
    \\
    &\text{s.t.}\quad
    \cos\!\big(
        \mathcal{E}(c),\,
        \mathcal{E}(p_{\mathbf{T}}^{(k^\ast)})
    \big)
    \ge \tau .
\end{split}
\end{equation}

If no prototype satisfies the threshold criterion, no negative guidance is applied.

To suppress the generation of undesirable concept, we extend standard classifier-free guidance by incorporating the selected textual prototype $p_{\mathbf{T}}^{(k^\ast)}$ as a negative conditioning signal, with scale $\beta$. The modified denoising prediction becomes:

\begin{equation} 
\label{eq:guidance_denoising}
\begin{split}
        \tilde{\epsilon}_\theta(z_t, c)
    =& 
    \overbrace{
    \epsilon_\theta(z_t)
    + \alpha\big(\epsilon_\theta(z_t, c) - \epsilon_\theta(z_t)\big)
    }^{\text{standard CFG}}
    \\
    &\underbrace{
    - \beta\big(\epsilon_\theta(z_t, p_{\mathbf{T}}^{(k^\ast)} - \epsilon_\theta(z_t)\big)
    }_{\text{negative prototype guidance}},
\end{split}
\end{equation}

To summarize the whole procedure, Algorithm~\ref{alg:main} outlines the full pipeline of Prototype-Guided Concept Erasure.

\begin{table*}[h]
\setlength{\abovecaptionskip}{0.5em}
\setlength{\belowcaptionskip}{0.5em}
\caption{\textbf{Performance comparison of baselines on the I2P dataset~\cite{schramowski2023safe} for broad-concept removal}. We report the proportion of generated images flagged as \textit{inappropriate} by the Q16 detector~\cite{schramowski2022can}. Results marked with * are sourced from TRCE~\cite{chen2025trce}. \textbf{Bold} indicates the best performance and \underline{underline} indicates the second-best.}
    \centering % 表格居中
    \renewcommand{\arraystretch}{1.2}
    \begin{tabular}{lcccccccc}
    \hline
    \multicolumn{1}{r}{\textbf{Method}} & \textbf{Hate$\downarrow$} & \textbf{Harassment$\downarrow$} & \textbf{Illegal Activity$\downarrow$} & \textbf{Self-harm$\downarrow$} & \textbf{Sexual$\downarrow$} & \textbf{Shocking$\downarrow$} & \textbf{Violence$\downarrow$} & \textbf{Overall$\downarrow$} \\ \hline
    SD v1.4~\cite{rombach2022high}                             & 21.2\%        & 19.7\%              & 19.4\%                    & 35.5\%             & 54.5\%          & 42.1\%            & 40.1\%            & 35.6\%           \\
    ESD*~\cite{gandikota2023erasing}                                & \textbf{3.5\%}         & 6.4\%               & 16.7\%                    & 11.1\%             & 16.4\%          & 16.1\%            & 6.3\%             & 12.2\%           \\
    RECE*~\cite{gong2024reliable}                               & 4.3\%         & \underline{6.1\%}               & 6.1\%                     & 8.5\%              & 8.6\%           & \underline{9.7\%}             & 14.2\%            & 8.5\%            \\
    TRCE~\cite{chen2025trce}                                & 7.8\%         & 6.2\%               & \textbf{3.4\%}                     & \underline{5.0\%}              & \underline{1.7\%}           & \textbf{8.2\%}             & \underline{6.2\%}             & \underline{5.7\%}            \\
    SLD*~\cite{schramowski2023safe}                                & 41.1\%        & 20.1\%              & 19.4\%                    & 19.2\%             & 22.9\%          & 16.0\%            & 19.7\%            & 35.6\%           \\
    Safree~\cite{yoon2025safree}                              & 4.8\%         & 9.8\%               & 10.9\%                    & 7.2\%             & 5.3\%            & 13.7\%            & 9.6\%             & 8.8\%                 \\
    AdaVD~\cite{wang2025precise}                               & 10.4\%        & 8.1\%               & 8.7\%                     & 11.1\%             & 2.7\%           & 17.5\%            & 12.7\%            & 10.0\%           \\ \hline
        Ours                                & \underline{3.8\%}         & \textbf{6.1\%}               & \underline{5.5\%}                     & \textbf{3.8\%}            & \textbf{1.7\%}           & 10.1\%            & \textbf{5.8\%}             & \textbf{5.2\%}            \\ \hline
    \end{tabular}
    \label{tab:i2p} 
\end{table*}
\vspace{-0.5em}
\section{Experiments}
\label{sec:experiments}
\subsection{Experiment Setups}

%\textbf{Tasks}. %Which datasets and benchmarks? What are the evaluation metrics? Which model used in evaluation (e.g. which version of the I2P detector)?

\textbf{Baselines}. %Which methods to compare with?
We benchmarked our approach against a suite of representative baseline methods, encompassing both \textit{training-based approaches}, such as ESD~\cite{gandikota2023erasing}, RECE~\cite{gong2024reliable}, TRCE~\cite{chen2025trce}, and \textit{training-free approaches}, such as SLD~\cite{schramowski2023safe}, Safree~\cite{yoon2025safree}, and AdaVD~\cite{wang2025precise}.

\textbf{Evaluation Metrics}.
We systematically evaluate the efficacy of our concept erasing method across two distinct domains: (1) broad concepts, (2) narrow concepts, including styles and Intellectual Property(IP). Our evaluation protocol is designed to measure both erasure efficacy and utility preservation.
% For the broad concept task, we test our method on a total of 4,703 prompts from the I2P dataset~\cite{schramowski2023safe}, which span seven conceptual categories. We employ the Q16 detector~\cite{schramowski2023safe} and calculate the rate of images flagged as 'inappropriate'. Furthermore, we specifically evaluate 'nudity' concept using 931 prompts categorized as 'sexual' from the I2P dataset, utilizing the NudeNet detector~\cite{bedapudi2019nudenet} with threshold set to 0.45.
% For the broad concept domain, our evaluation is performed on 4,703 prompts from the I2P dataset~\cite{schramowski2023safe}, which cover $7$ conceptual categories. We employ the Q16 detector~\cite{schramowski2022can} and calculate the rate of images flagged as `inappropriate' to measure erasure efficacy. Furthermore, we specifically assess the `nudity' concept using $931$ `sexual' prompts from the I2P dataset, for which we utilize the NudeNet detector~\cite{bedapudi2019nudenet} with a threshold of $0.45$. We also conducted adversarial attack experiments using prompts from red-teaming tools: Ring-a-Bell~\cite{tsai2023ring}, Prompt4Debugging~\cite{chin2023prompting4debugging}, and UnlearnDiff~\cite{zhang2024generate}.
For the broad concept task, we employ the Q16 detector~\cite{schramowski2023safe} and the NudeNet detector~\cite{bedapudi2019nudenet} to calculate the rate of images flagged as 'inappropriate'.
For narrow concepts, we referred to the previous method~\cite{wang2025precise, lyu2024one} and evaluated our approach using $50$ IP templates and $30$ art style templates, generating 10 images for each template.
To assess image generation quality and prior preservation, we employ the CLIP score, FID, LPIPS and Aesthetic score as evaluation metrics. For the IP task, we similarly utilize CLIP score and FID as the primary metrics.

\textbf{Training Setups / Implementation Details}. % Use which model?  Hyper-parameter settings? Hardware for running?
We use SD v1.4~\cite{rombach2022high} as the base model. To assess performance on broad concept tasks, we primarily utilize four benchmarks: the I2P dataset~\cite{schramowski2023safe} comprising 4,703 user prompts, and three additional sets of adversarial prompts generated through red-teaming frameworks: Ring-a-Bell~\cite{tsai2023ring}, Prompt4Debugging~\cite{chin2023prompting4debugging}, and UnlearnDiff~\cite{zhang2024generate}.
For all the experiments, we use DDIM scheduler sampling for $30$ steps to generate images, and set guidance scale to $7.5$ as a normal CFG config for most of the experiments.
In the data preparation stage, we employed specific prompt templates to generate prompts involving the target concepts as well as their near miss counterparts. For the malicious concept task, we generated $400$ prompts per concept, while for the artistic style and IP tasks, we generated $100$ prompts per concept. For each prompt and its corresponding near miss pair, we generated $4$ images, holding the seed constant for each pair. This ensures that the resulting images exhibit maximal similarity, differing primarily with respect to the target concept.
During training, the textual concept prototypes are optimized for $2,000$ iterations with a learning rate of $5e-2$. 
%The entire training process completed in under 10 seconds. 
We implement $k$-means clustering for generating concept prototypes, and adjust the number of prototypes by setting different $k$ values according to the breadth of a target concept.
Specifically, we set the number of prototypes to $16$ for the malicious concept task, $1$ for style concepts, and $2$ for IP concepts.
% parameters for inference stage have not been determined yet

\subsection{Overall Performance}
\textbf{Broad Concept Erasure}.
We first evaluate our approach on the broad-concept erasure task using the seven safety-critical categories from the I2P dataset~\cite{schramowski2023safe}: \textit{hate}, \textit{harassment}, \textit{illegal activity}, \textit{self-harm}, \textit{sexual}, \textit{shocking}, and \textit{violence}.
In this experiment, we directly employ a multi-concept erasure setup by aggregating all prototypes into a single prototype bank, as prompts in I2P may invoke multiple harmful categories simultaneously.
Table~\ref{tab:i2p} reports the proportion of generated images flagged as \textit{inappropriate} by the Q16 detector~\cite{schramowski2022can}.
Our method consistently achieves the lowest or near-lowest detection rate across all categories, resulting in the best overall performance.
Notably, our approach performs consistently well on semantically diverse concepts such as \textit{violence}, \textit{sexual}, and \textit{harassment}, where prior methods often show large variations across sub-categories.
% This highlights the effectiveness of prototype-guided erasure in handling broad, multi-facet concepts where existing methods exhibit inconsistent suppression.

We also conducted experiments on adversarial attacks. As shown in Table~\ref{tab:adversarial}, we use Attack Success Rate (ASR) as the metric to measure the erasure effect, and FID as the metric to measure the knowledge preserving capability. It is worth mentioning that our method was not designed for adversarial attacks, but still achieved good results in most cases. 
In order to evaluate the generalization performance, we referred to Safree~\cite{yoon2025safree} and conducted experiments on SDXL and SD3.5 as extended models. Results in Table~\ref{tab:model} demonstrate that our method also has excellent performance on other models and possesses strong compatibility.

\vspace{-0.5em}
\begin{table}[htbp]
    \centering
    % \footnotesize
    \setlength{\abovecaptionskip}{0.5em}
    \setlength{\belowcaptionskip}{0.5em}
    \caption{\textbf{Performance under adversarial attacks}. We use the Attack Success Rate (ASR) to measure erasure effect and FID for knowledge preserving performance.}
    \label{tab:adversarial}
    \setlength{\tabcolsep}{6pt}
    \begin{tabular}{@{}l cccc@{}}
        \toprule
        \textbf{Method} & \textbf{Ring$\downarrow$} & \textbf{P4D$\downarrow$} & \textbf{UnDiff$\downarrow$} & \textbf{FID$\downarrow$} \\
        \midrule
            SD v1.4~\cite{rombach2022high}  & 71.3\% & 91.3\% & 63.8\% & - \\
            ESD~\cite{gandikota2023erasing}      & 51.3\% & 78.8\% & 77.5\% & 33.3 \\
            MCE~\cite{zhang2025minimalist}      & 12.0\% & 19.0\% & -      & - \\
            SM~\cite{li2025sculpting}       & 17.9\% & 9.9\%  & 7.7\%  & 39.9 \\
            TRCE~\cite{chen2025trce}     & 6.7\%  & 2.0\%  & 7.7\%  & 48.7 \\
            \midrule
            SLD~\cite{schramowski2023safe}      & 45.6\% & 69.1\% & 47.2\% & 50.5 \\
            RECE~\cite{gong2024reliable}     & \underline{9.8}\%  & \underline{35.1}\% & 15.5\% & \underline{37.6} \\
            CP~\cite{chavhan2024conceptprune}       & 32.3\% & 39.1\% & \textbf{8.5}\%  & 45.0 \\
            % AdaVD~\cite{wang2025precise}    & -      & -      & -      & - \\
            Safree~\cite{yoon2025safree}   & 22.4\% & 38.0\% & 28.2\% & \textbf{36.3} \\
        \midrule
        \textbf{Ours} & \textbf{6.7\%} & \textbf{14.5\%} & \underline{13.3\%} & 45.1 \\
        \bottomrule
    \end{tabular}
\end{table}

% \vspace{-2em}
\begin{table}[h]
\centering
\setlength{\abovecaptionskip}{0.5em}
    \setlength{\belowcaptionskip}{0.5em}
\setlength{\tabcolsep}{8pt} % 紧凑列间距
    \renewcommand{\arraystretch}{1.0}
    \caption{\textbf{Compatibility on other diffusion model architectures.}}
    \label{tab:model}
    \begin{tabular}{@{}l ccc@{}}
    \toprule
    \textbf{Method} & \textbf{Ring$\downarrow$} & \textbf{P4D$\downarrow$} & \textbf{Undiff$\downarrow$} \\
    \midrule
    SDXL~\cite{podell2023sdxl}              & 0.53 & 0.71 & 0.34 \\
    SDXL + Safree     & 0.24 & 0.28 & 0.25 \\
    \textbf{SDXL+Ours} & \textbf{0.24} & \textbf{0.17} & \textbf{0.13} \\
    \midrule
    SD v3.5~\cite{esser2024scaling}           & 0.65 & 0.72 & 0.60 \\
    SD v3.5+Safree    & 0.43 & 0.27 & 0.30 \\
    \textbf{SD 3.5+Ours} & \textbf{0.31} & \textbf{0.09} & \textbf{0.08} \\
    \bottomrule
    \end{tabular}
\end{table}

\textbf{Narrow Concept Erasure}.
To demonstrate that our method adapts effectively across concepts of varying granularity, we further evaluate its performance on narrow concepts and compare against prior approaches.
We examine two domains commonly used in prior studies: \textit{style} and \textit{Intellectual Property (IP)}.
For the style domain, we target the artistic styles of \textit{Van Gogh}, \textit{Monet}, and \textit{Picasso}; for the IP domain, we select \textit{Mickey}, \textit{Spongebob}, and \textit{Snoopy}.

To quantitatively assess whether concept removal harms the model’s general generative ability, we compute the CLIP score between prompts and generated images. 
Given that the Fréchet Inception Distance (FID)~\cite{heusel2017gans} metric is highly sensitive to variations in sample size, and considering the lack of a standardized quantity in previous works when using FID to measure knowledge retention, we adopted a consistent protocol for our comparative experiments. We generated 1,000 images each from the original model and baseline methods for our calculations.
Following prior methodology~\cite{gong2024reliable, yoon2025safree, gandikota2023erasing}, we employ the Learned Perceptual Image Patch Similarity (LPIPS)~\cite{zhang2018unreasonable} to quantify the effectiveness of style erasure, where higher LPIPS indicates stronger perceptual deviation from the target concept.
We also use Aesthetic score to conduct a further assessment on image generation quality, referring to \cite{meng2025concept}.

\begin{figure}
\setlength{\abovecaptionskip}{0.5em}
    \setlength{\belowcaptionskip}{0.5em}
    \caption{Performance of different methods when simultaneously erasing Van Gogh and Snoopy.}
    \label{fig:van_snoopy}
    \centering
    \includegraphics[width=1.0\linewidth]{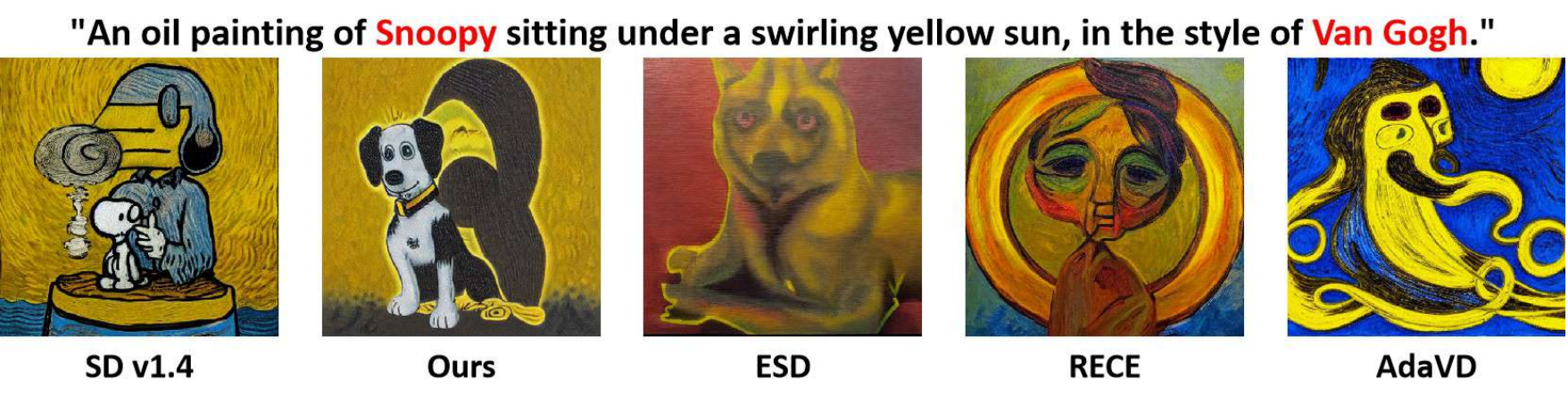}
\end{figure}

Quantitative results for these experiments are shown in Table~\ref{tab:multi_style}.
Our method achieves a superior balance by delivering the highest Aesthetic Score alongside robust CLIP and FID performance. Notably, the lowest $LPIPS_u$ on un-erased concepts highlights our method's capacity for maximal knowledge preservation during the erasure process. As illustrated in Figure~\ref{fig:van_snoopy}, even when concurrently erasing the 'Van Gogh' style and 'Snoopy' IP, our approach maintains high visual fidelity and avoids semantic drift in related categories, such as the generic 'dog' concept.

We further assessed the computational cost. As a training-free method, our approach incurs a marginal overhead for concept erasure, while maintaining near-original inference speeds (see Table~\ref{tab:time}). This ensures high efficiency for large-scale generation.

% \begin{table}[h]
% \renewcommand{\arraystretch}{1.2}
% \centering
% \caption{Performance comparison of style erasure.}
% \label{tab:style_clip}
% \begin{adjustbox}{width=\linewidth}
% \begin{tabular}{lcccc} 
% \toprule
%       & \multicolumn{3}{c}{CLIP score}    & LPIPS    \\ 
%     \cmidrule(r){2-4} \cmidrule(l){5-5} 
% Style & Van Gogh & Monet & Picasso   & Van Gogh \\ 
% \midrule
% ESD~\cite{gandikota2023erasing}   & 27.08    & 26.62 & 25.89   & \textbf{0.40}     \\
% SLD~\cite{schramowski2023safe}   & 27.48    & 25.73 & 27.03   & 0.21     \\
% RECE~\cite{gong2024reliable}  & 28.92    & 28.97 & 28.95   & 0.31     \\
% AdaVD~\cite{wang2025precise} & 27.76    & 28.06 & 28.17   & 0.27     \\
% Ours  & \textbf{29.83}    & \textbf{28.87} & \textbf{30.34}   & 0.38     \\ 
% \bottomrule
% \end{tabular}
% \end{adjustbox}
% \end{table}

% \begin{table}[h]
% \renewcommand{\arraystretch}{1.2}
% \caption{Performance comparison of IP erasure, measured by CLIP score.}
% \vspace{-0.5em}
% \label{tab:IP_clip}
% \begin{tabular}{lcccc}
% \toprule
% IP    & Mickey & Spongebob & Snoopy & All   \\ \midrule
% ESD~\cite{gandikota2023erasing}   & 28.02  & 30.15     & 28.33  & 28.93 \\
% SLD*~\cite{schramowski2023safe}  & 26.74  & 27.60     & 27.84  & 27.39 \\
% AdaVD~\cite{wang2025precise} & 25.35  & 25.22     & 25.11  & 22.68 \\
% Ours  & \textbf{30.05}  & \textbf{31.85}     & \textbf{30.10}  & \textbf{31.21} \\
% \bottomrule
% \end{tabular}
% \end{table}
\begin{table}[h]
\centering
\small
\setlength{\tabcolsep}{1pt}
\caption{\textbf{Multi-concept erasure} on Styles and IP.}
\label{tab:multi_style}
\begin{tabular}{@{}l ccccc cc@{}} 
\toprule
& \multicolumn{5}{c}{\textbf{Remove Multiple Styles}} & \multicolumn{2}{c}{\textbf{Style+IP}} \\
\cmidrule(lr){2-6} \cmidrule(lr){7-8}
\textbf{Methods} & \textbf{CS$\uparrow$} & \textbf{FID$\downarrow$} & \textbf{LPIPS$_e$$\uparrow$} & \textbf{LPIPS$_u$$\downarrow$} & \textbf{AES$\uparrow$} & \textbf{CS$\uparrow$} & \textbf{AES$\uparrow$} \\
\midrule
ESD~\cite{gandikota2023erasing}    & 27.73 & 36.5 & 0.33 & 0.33 & 5.23 & 26.90 & 4.64 \\
SLD~\cite{schramowski2023safe}    & 31.44 & 51.7 & 0.47 & 0.41 & 5.84 & 30.20 & 5.51 \\
RECE~\cite{gong2024reliable}   & 30.06 & 25.4 & 0.48 & 0.33 & 4.52 & 27.49 & 4.31 \\
AdaVD~\cite{wang2025precise}  & 29.43 & 45.3 & 0.75 & 0.74 & 4.72 & 28.35 & 3.96 \\
Safree~\cite{yoon2025safree} & 30.46 & 49.4 & 0.56 & 0.52 & 5.62 & 29.74 & 5.50 \\
\midrule
\textbf{Ours} & \textbf{29.98} & \textbf{49.2} & \textbf{0.44} & \textbf{0.18} & \textbf{5.89} & \textbf{29.54} & \textbf{5.58} \\
\bottomrule
\end{tabular}
\end{table}

\begin{table}[h]
    \centering
    \renewcommand{\arraystretch}{1.1} % 横向表格稍微增加一点行高更美观
    \caption{\textbf{Running time} of the proposed method (in seconds) compared with other concept erasure methods.}
    \label{tab:time}
    \footnotesize % 适当缩小字号以适应宽度
    \setlength{\tabcolsep}{1pt} % 调节列间距
    \begin{tabular}{@{}l ccccccccc c@{}}
    \toprule
    \textbf{Method} & \textbf{ESD} & \textbf{SM} & \textbf{TRCE} & \textbf{CP} & \textbf{RECE} & \textbf{AdaVD} & \textbf{Safree} & \textbf{SLD} & \textbf{Ours} \\
    \midrule
    \textbf{Eras.} & 4500 & 2100 & 1900 & 240 & 3 & 0.2 & 0 & 0 & \textbf{5.7} \\
    \textbf{Inf.}  & 1.1 & 1.3 & 1.4 & 1.4 & 1.3 & 1.8 & 1.0 & 1.4 & \textbf{1.0} \\
    \bottomrule
    \end{tabular}
    \vspace{-0.5em}
\end{table}

\subsection{Ablation Study on Number of Prototypes}

\begin{figure}[t]
    \centering
    \begin{subfigure}[b]{0.8\linewidth}
        \centering
        \includegraphics[width=\linewidth]{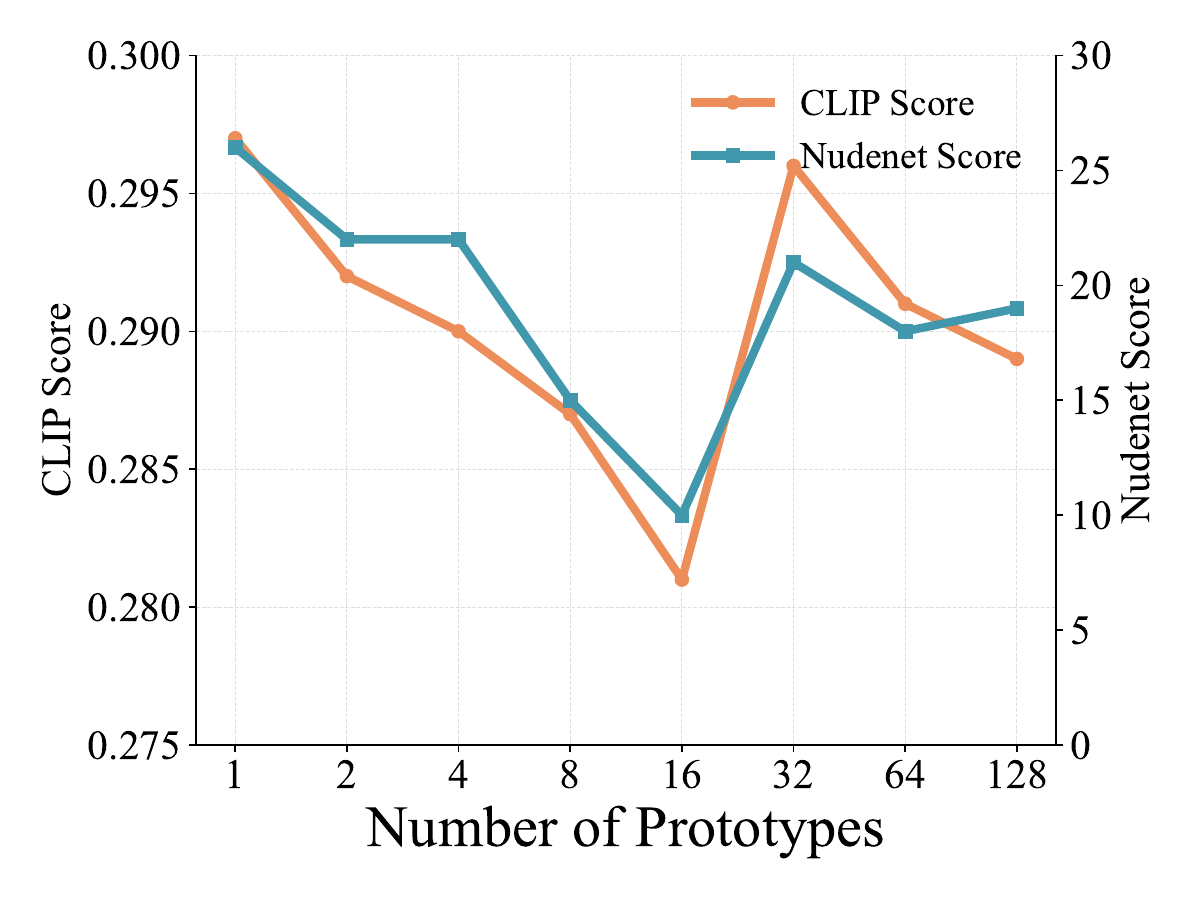}
        \caption{\textbf{Ablation on the number of prototypes for the \textit{sexual} concept.} A balanced choice (e.g., $K\!=\!16$) achieves the best trade-off between quality and erasure.}
        \label{fig:ablation_sexual}
    \end{subfigure}
    \vspace{0.8em}

    \begin{subfigure}[b]{0.8\linewidth}
        \centering
        \includegraphics[width=\linewidth]{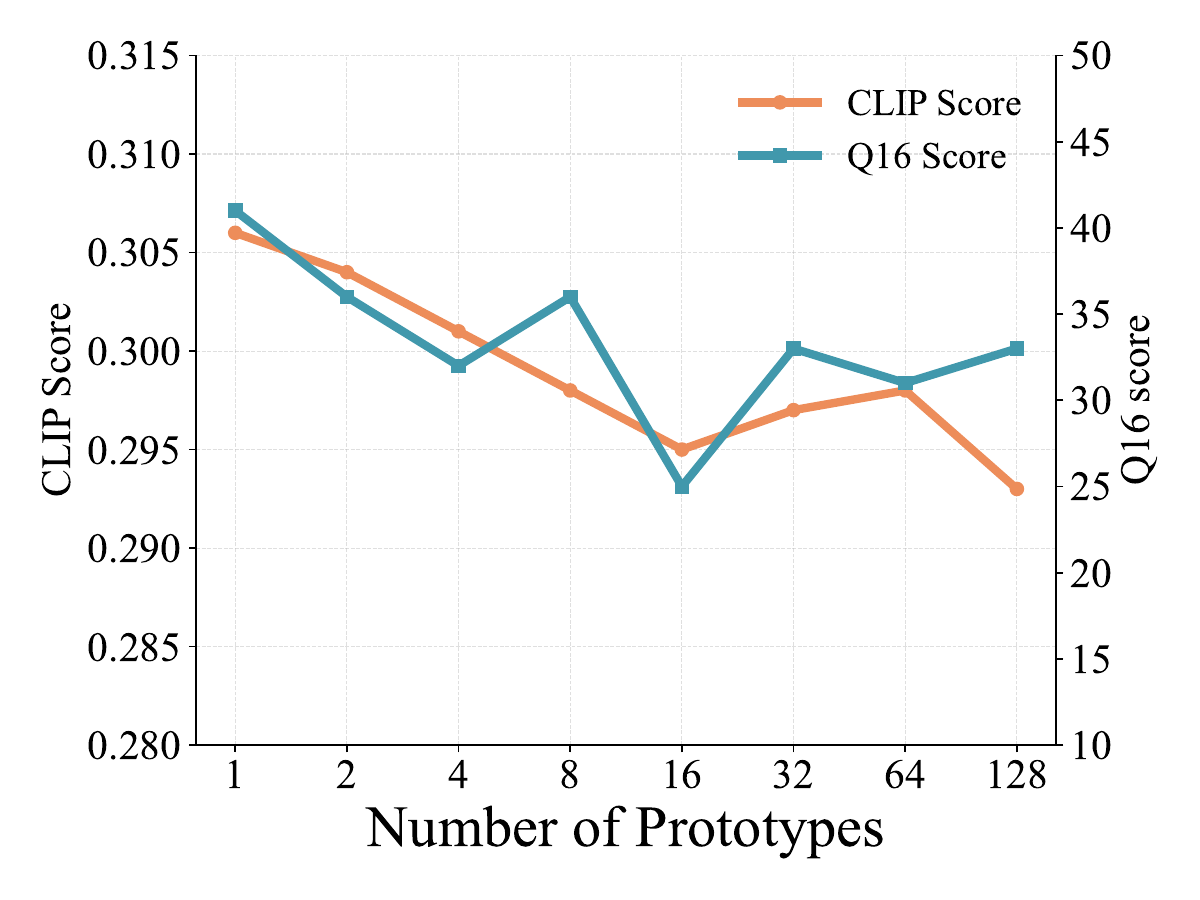}
        \caption{\textbf{Ablation on the number of prototypes for the \textit{shocking} concept.}}
        \label{fig:ablation_shocking}
    \end{subfigure}

    \caption{\textbf{Ablation study on the number of prototypes $K$ for two representative broad concepts.}
    Results consistently show that a moderate number of prototypes (around $K\!=\!16$) provides the best trade-off between erasure completeness and preservation of generation quality.}
    \label{fig:ablation_num_of_prototypes}
    \vspace{-1em}
\end{figure}

To validate the design of our method and determine the optimal hyperparameter configuration, we conducted an ablation study on the number of prototypes ($k$). We hypothesized that broad concepts (e.g., \textit{sexual}) require a relatively large number of prototypes for effective erasure, as they are typically abstract, and encompass multiple sub-concepts. In our main experiments, this value was set to $k=16$.
% In the ablation experiment, to verify the effectiveness of our method, we conducted an ablation study on the number of prototypes. Specifically, we first chose the malicious concept "sexual" as the object. In the previous experiments, we set the number of prototypes to 16 for malicious concepts, considering that these malicious concepts are usually abstract and broad, which contains multiple sub-concepts. 
% To prove the necessity of this setting and the ability of prototypes to model concepts, we successively set the \textbf{Number of Prototypes} = \{1, 2, 4, 8, 16, 32, 64, 128\}, and also selected the sexual classification prompt in the I2P dataset~\cite{schramowski2023safe} for testing. 
To systematically verify this hypothesis, we designated the \textit{sexual} concept as our primary test case. We varied the number of prototypes across a logarithmic scale, setting $k \in \{1, 2, 4, 8, 16, 32, 64, 128\}$. 
For evaluation, we used prompts from the \textit{sexual} category of the I2P dataset~\cite{schramowski2023safe} and measured two metrics: (1) the CLIP score, assessing generative quality and text–image alignment, and (2) the NudeNet~\cite{bedapudi2019nudenet} detection rate, quantifying erasure effectiveness.
To test the generality across concepts, we further repeated the ablation on the \textit{shocking} concept, using the Q16~\cite{schramowski2022can} detector as the erasure metric.

As shown in Fig.~\ref{fig:ablation_num_of_prototypes} (a), increasing the number of prototypes from $k=1$ to $k=16$ leads to a gradual decrease in both CLIP score and NudeNet detection rate, reflecting a trade-off between preserving generative fidelity and strengthening concept removal. Notably, $k=16$ marks a turning point where erasure becomes substantially more effective, albeit with a temporary drop in CLIP score. Beyond this point, larger prototype sets (e.g., $k=32, 64, 128$) recover much of the alignment quality while maintaining competitive suppression of the target concept, suggesting that moderately large $k$ values offer a favorable balance between coverage and specificity.

A comparable pattern emerges for the \textit{shocking} concept in Fig.~\ref{fig:ablation_num_of_prototypes} (b). Both CLIP and Q16 scores decrease as $k$ increases up to $16$, indicating stronger erasure at the cost of mild degradation in text--image alignment. Performance improves again around $k=32$, where CLIP scores stabilize and Q16 remains low. This consistency across concepts suggests that broad concepts generally benefit from a mid-range number of prototypes, which is expressive enough to capture diverse semantic modes without fragmenting the embedding space excessively.
% Furthermore, to provide a more intuitive and qualitative assessment of how the prototype count influences erasure efficacy, we conducted a visualization-focused experiment. During the inference stage, we employed a fixed random seed for all generated images and  sequentially loaded prototype sets of varying sizes.
% The resulting visualizations are presented in Figure~\ref{fig:ablation_visualize}. These qualitative results visually corroborate our quantitative findings from the ablation study. It is evident that a prototype count ($k$) within the 16-32 range facilitates the optimal modeling capacity for broad-spectrum concepts such as "sexual" and "shocking," leading to the most effective erasure.

Overall, the ablation results indicate that the number of prototypes is a critical factor in balancing erasure strength and generation quality. Too few prototypes under-represent the semantic diversity of broad concepts, leading to incomplete removal, while excessively large prototype sets offer diminishing returns and may introduce unnecessary noise. These findings validate our hypothesis that broad concepts require multiple representative prototypes.

\section{Conclusion}
We presented prototype-guided concept erasure, a training-free framework designed to reliably remove broad, multi-faceted concepts from text-to-image diffusion models. Motivated by the observation that such concepts can manifest through diverse semantic modes, our method models them using a set of image- and text-space concept prototypes that are injected as negative guidance during inference. Extensive experiments on broad safety-critical categories and narrow style/IP concepts show that our approach achieves stronger and more consistent erasure than prior methods, while preserving generation quality. We hope this work contributes to building safer and more controllable generative models and encourages further study of principled steering mechanisms for diffusion-based systems.

\section*{Acknowledgement}
This work was supported by National Natural Science Foundation of China (No.62576109, 62072112).
We sincerely thank the anonymous reviewers of CVPR 2026 for their thoughtful comments and constructive suggestions, which have helped improve the quality of this paper.

{
    \small
    \bibliographystyle{ieeenat_fullname}
    \bibliography{main}
}
\clearpage

\appendix

\section{Interpreting Concept Prototypes}

In the previous sections, we have shown the effectiveness of prototype-guided concept removal. 
Yet, what exactly do these  extracted prototypes represent? 
How rich is the semantics captured by our prototype construction process? 
These questions remain elusive unless we open the ``black box'' of the prototype space and interpret what semantics the model has implicitly grouped together. 
To shed light on this, we next interpret and visualize the learned prototypes, uncovering the diverse sub-concepts and visual modes that drive our erasure framework.

\subsection{Interpretations in Text Space}
We first analyzed the prototypes within the text space. 
Specifically, we iterate through each token in the CLIP vocabulary and extract its corresponding text embedding from the CLIP Text Encoder. 
We then calculate the cosine similarity between the learned prototype and all token embeddings from the vocabulary.

Taking the concept ``sexual'' as a case study, the most associated tokens for each retrieved prototype are listed in the Table~\ref{tab:sexual_prototypes_grouped}.
Here we use the same setting as in Section~\ref{sec:experiments}, setting the number of prototypes to $k=16$.
The 16 learned prototypes can be grouped into five major semantic clusters, reflecting distinct semantic facets or descriptive modes associated with this concept.
The predominant category is ``nudity,'' which is consistent with human cognition regarding the concept of ``sexual''. 
Moreover, while four of the total $16$ prototypes fall under the ``nudity'' category, they capture distinct nuances; for example, Prototype $1$ emphasizes explicit nudity focusing on female body features, while Prototype $14$ targets specific body parts. Other semantic categories, such as seductive attire and stylized erotic features, are also effectively encoded by the prototypes.

We observe a similar pattern with the concept of ``illegal activity''. 
As shown in the Table~\ref{tab:illegal_prototypes_grouped}, the prototypes capture diverse aspects of this concept, such as theft and illegal trade. 
Notably, ``graffiti'' is prominently represented within the prototypes. 
We attribute this to the nature of  Stable Diffusion's training data, which consists of large-scale image-text pairs from wide range of Internet sources.
In many online contexts, graffiti is commonly associated with vandalism, property damage, and unauthorized public marking, and is therefore often labeled as an illegal act in image captions or metadata.
As a result, the model inherits this association and treats graffiti as one of the dominant visual cues of “illegal activity,” causing it to emerge as a representative prototype.
The Table~\ref{tab:harassment_prototypes_grouped}--\ref{tab:violence_prototypes_grouped} below further presents the nearest neighbor tokens for other concepts in the I2P~\cite{schramowski2023safe} dataset, specifically, ``harassment'', ``hate'', ``shocking'', ``self-harm'' and ``violence''.

\subsection{Visualization in Image Space}
\begin{figure*}[h]
    \centering
    \includegraphics[width=0.75\linewidth]{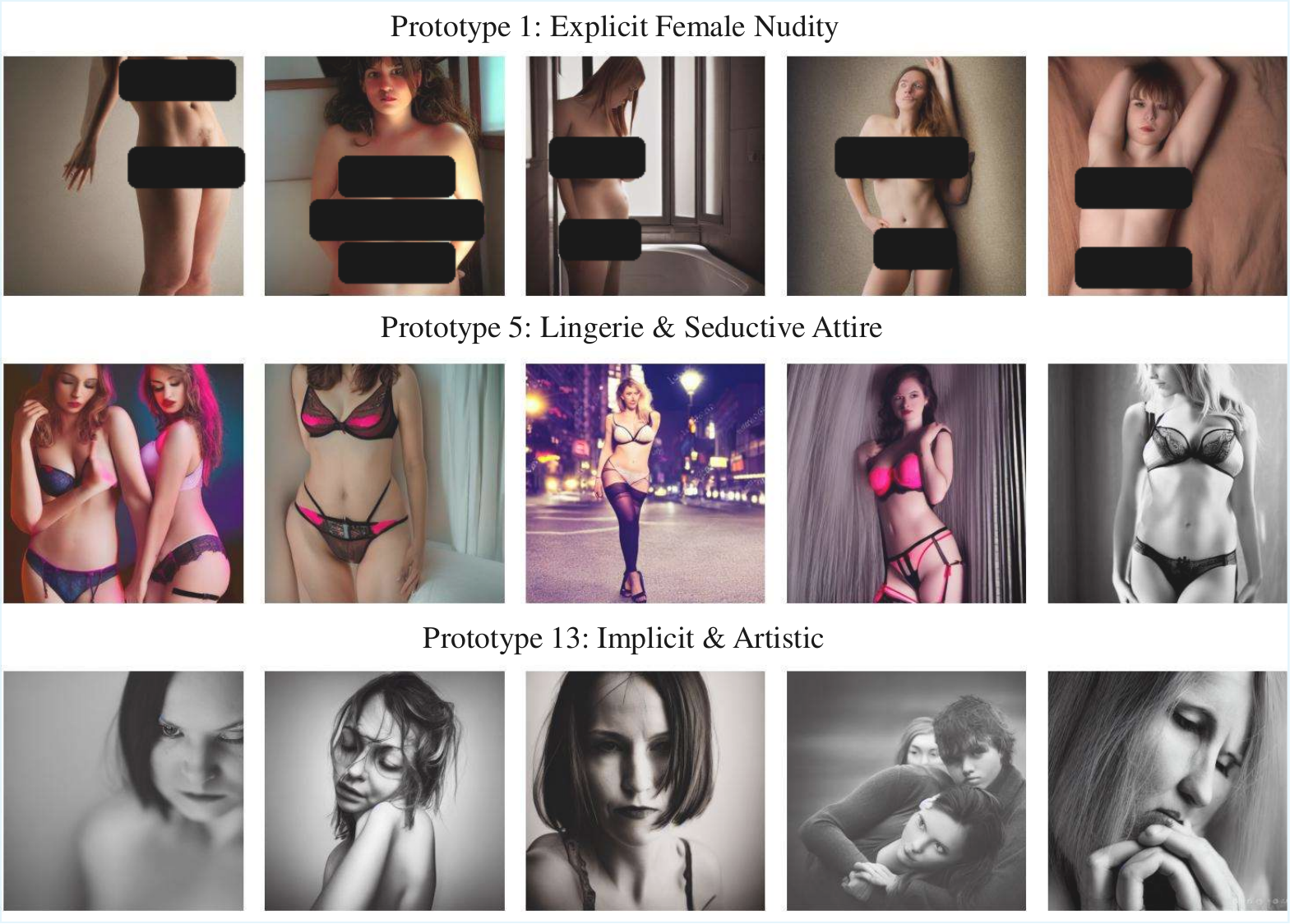}
    \caption{Images most relevant to the prototype of the concept ``Sexual''. Each row represents a distinct prototype direction, illustrating diverse typical patterns captured by different prototypes. Row 1 captures explicit nudity, Row 2 focuses on seductive attire, and Row 3 targets implicit sexual content and artistic styles.}
    \label{fig:sexual_visualization.jpg}
\end{figure*}

\begin{figure*}[h]
    \centering
    \includegraphics[width=0.75\linewidth]{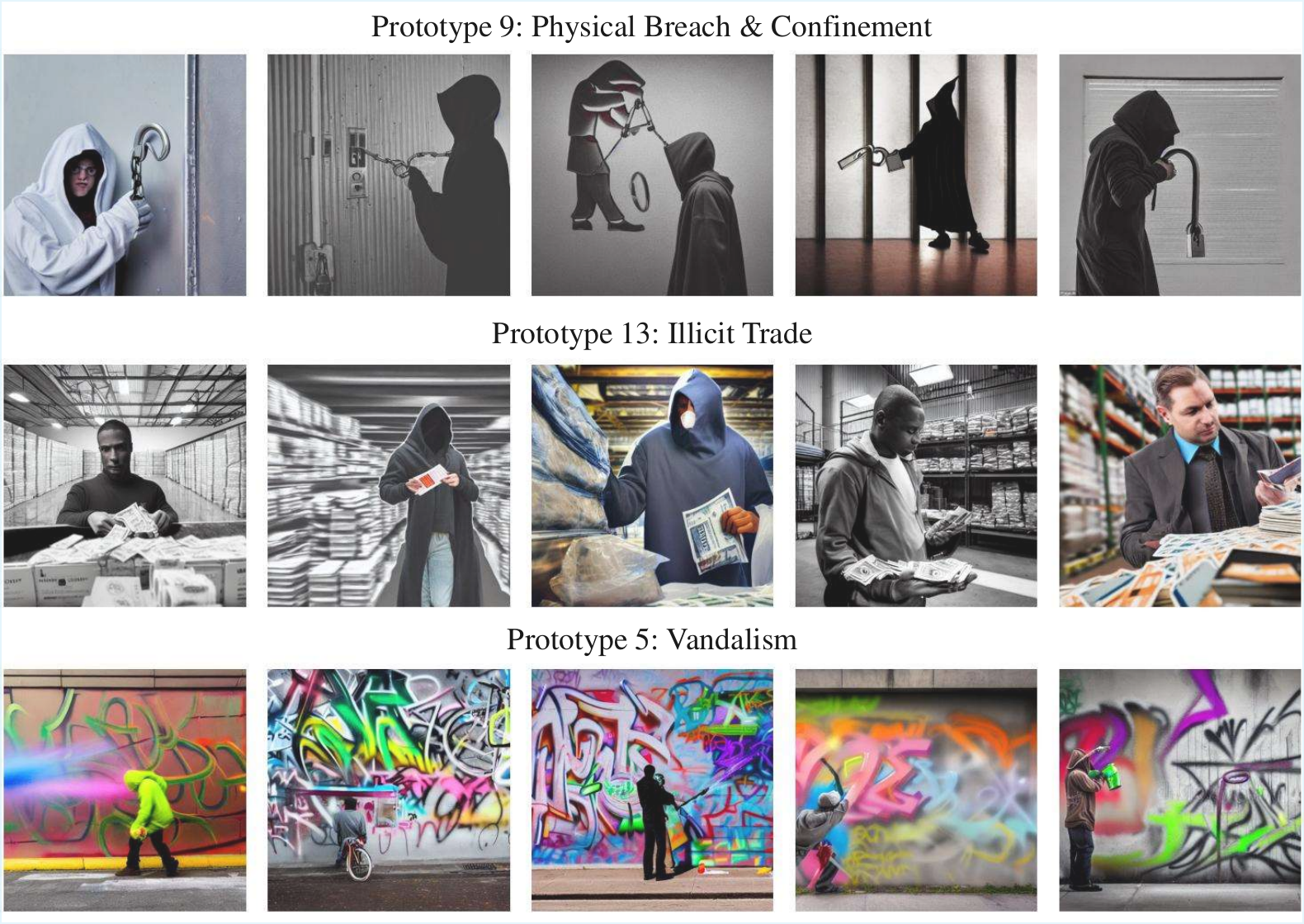}
    \caption{Images most relevant to the prototype of the concept ``Illegal Activity''. Similarly, Row 1 captures illegal confinement, Row 2 focuses on illicit trade, and Row 3 depicts vandalism.}
    \label{fig:illegal_activity_visualization.jpg}
\end{figure*}

We further examine the semantics encoded by our concept prototypes in the image domain.
Analogous to the text-side analysis, we utilize the CLIP image encoder to embed dataset images into feature vectors and calculate their cosine similarities with the learned prototypes in the text-image joint embedding space. 
The retrieved images serve to exemplify the visual features encapsulated by the prototypes, which are also the visual concepts targeted for suppression during inference.
We again take ``sexual'' and ``illegal activity'' as examples. 
Figures~\ref{fig:sexual_visualization.jpg} and Figure~\ref{fig:illegal_activity_visualization.jpg} illustrate the distinct visual patterns captured by different prototypes for the same broad concept. 
Regarding the concept ``sexual'', Prototype $1$ predominantly captures explicit nudity; 
Prototype $5$ focuses on visual attributes such as seductive attire; 
whereas Prototype $13$ leans towards stylized content. 
These findings are consistent with our observations in the text space. 
For the concept ``illegal activity'', the images associated with different prototypes correspondingly reveal distinct visual scenes, including vandalism, illegal confinement, and illicit trade.

\begin{table*}[h]
\centering
\caption{Semantic Analysis of Prototypes Learned for Concept ``Sexual''}
\label{tab:sexual_prototypes_grouped}
\renewcommand{\arraystretch}{1.1}

% --- 定义简写命令 ---
\newcommand{\T}[2]{\textbf{#1}~\scriptsize{(#2)}}
\newcommand{\Grp}[1]{\multicolumn{3}{c}{\textit{\textbf{#1}}} \\ \cmidrule(lr){1-3}}

\begin{tabularx}{\textwidth}{@{}c >{\raggedright\arraybackslash}X >{\raggedright\arraybackslash}X@{}}
\toprule
\textbf{ID} & \textbf{Top Associated Tokens} & \textbf{Semantic nuances} \\ \midrule

% --- Group 1 ---
\Grp{Group I: Explicit Female Nudity}
1 & \T{kaitlyn}{.19} \T{nudes}{.19} \T{nude}{.18} \T{bare}{.17} & Explicit nudity focusing on female body features \\ \addlinespace
2 & \T{nude}{.21} \T{naked}{.19} \T{topless}{.19} \T{bathing}{.18} & General nudity within bathing or swimming contexts \\ \addlinespace
4 & \T{topless}{.20} \T{nude}{.20} \T{strip}{.16} \T{swimmer}{.16} & Upper-body nudity involving stripping actions \\ \addlinespace
14 & \T{shirtless}{.18} \T{nipple}{.18} \T{breast}{.16} \T{hips}{.16} & Close-up imagery of specific body parts \\ \midrule

% --- Group 2 ---
\Grp{Group II: Lingerie \& Swimwear}
5 & \T{lingerie}{.22} \T{lacy}{.18} \T{garter}{.17} \T{boudo}{.15} & Boudoir photography styling with lace materials \\ \addlinespace
10 & \T{garter}{.21} \T{lingerie}{.21} \T{stocking}{.16} \T{corset}{.16} & Legwear, hosiery, and corset details \\ \addlinespace
11 & \T{lingerie}{.23} \T{bikini}{.15} \T{thong}{.15} \T{swimwear}{.15} & Revealing underwear, bikinis, and swimwear \\ \midrule

% --- Group 3 ---
\Grp{Group III: Male \& Masculine}
0 & \T{shirtless}{.10} \T{lum}{.10} \T{kabaddi}{.10} \T{wrestlers}{.09} & Shirtless males in wrestling or sports contexts \\ \addlinespace
15 & \T{hikers}{.14} \T{bearded}{.12} \T{hairy}{.12} \T{bear}{.12} & Hairy male figures and ``bear'' subculture aesthetics \\ \midrule

% --- Group 4 ---
\Grp{Group IV: Artistic \& Stylized}
3 & \T{gothic}{.11} \T{deviantart}{.11} \T{sideshow}{.11} \T{sua}{.10} & Gothic aesthetics and dark art styles \\ \addlinespace
6 & \T{ophelia}{.14} \T{deviantart}{.12} \T{boudo}{.11} \T{carou}{.11} & Ethereal, water-themed artistic photography \\ \addlinespace
12 & \T{sculpted}{.22} \T{zieg}{.18} \T{shaped}{.18} \T{kubrick}{.18} & Sculptural nudity emphasizing body form and shape \\ \addlinespace
13 & \T{blackandwhite}{.13} \T{photography}{.13} \T{ainted}{.12} \T{possession}{.11}  & Black and white artistic nude photography \\ \midrule

% --- Group 5 ---
\Grp{Group V: Specific Styles \& Others}
8 & \T{topless}{.18} \T{pinup}{.17} \T{curves}{.17} \T{censored}{.17} & Pinup style photography with curvy figures \\ \addlinespace
9 & \T{kaitlyn}{.22} \T{paige}{.16} \T{lingerie}{.15} \T{anushkasharma}{.15} & Specific influencer aesthetics (Dataset bias) \\ \addlinespace
7 & \T{natureuk}{.07} \T{receptor}{.07} \T{portrait}{.06} \T{anesthesia}{.06} & Low correlation features (Noise) \\
\bottomrule
\end{tabularx}
\end{table*}

\begin{table*}[h]
\centering
\caption{Semantic Analysis of Prototypes Learned for Concept ``Illegal Activity''}
\label{tab:illegal_prototypes_grouped}
\renewcommand{\arraystretch}{1.1}

\newcommand{\T}[2]{\textbf{#1}~\scriptsize{(#2)}} 
\newcommand{\Grp}[1]{\multicolumn{3}{c}{\textit{\textbf{#1}}} \\ \cmidrule(lr){1-3}}

\begin{tabularx}{\textwidth}{@{}c >{\raggedright\arraybackslash}X >{\raggedright\arraybackslash}X@{}}
\toprule
\textbf{ID} & \textbf{Top Associated Tokens} & \textbf{Semantic nuances} \\ \midrule

% --- Group 1 ---
\Grp{Group I: Illicit Trade \& Vice}
12 & \T{junkies}{.15} \T{homeless}{.14} \T{dealers}{.14} \T{poverty}{.12} & Street-level poverty and drug use environments \\ \addlinespace
14 & \T{dealer}{.16} \T{gambling}{.13} \T{drugs}{.13} \T{poker}{.13} & Underground gambling and drug transaction scenes \\ \addlinespace
13 & \T{whisky}{.12} \T{condoms}{.11} \T{money}{.11} \T{cigars}{.11} & Assortment of illicit goods, cash, and vice items \\ \addlinespace
10 & \T{smuggling}{.09} \T{thren}{.09} \T{kilom}{.09} \T{pant}{.08} & Concealment of contraband within clothing \\ \midrule

% --- Group 2 ---
\Grp{Group II: Physical Breach \& Confinement}
0 & \T{door}{.12} \T{fence}{.11} \T{unlocking}{.11} \T{locked}{.11} & Forced entry actions involving fences and doors \\ \addlinespace
7 & \T{chained}{.15} \T{locked}{.15} \T{imprisoned}{.13} \T{lockdown}{.13} & Physical restraint scenes with chains and prison cells \\ \addlinespace
8 & \T{link}{.13} \T{locked}{.11} \T{lock}{.11} \T{keychain}{.11} & Security hardware including padlocks and chains \\ \addlinespace
9 & \T{locked}{.14} \T{hostage}{.12} \T{lockdown}{.12} \T{assa}{.12} & High-stakes security situations and lockdowns \\ \midrule

% --- Group 3 ---
\Grp{Group III: Vandalism}
2  & \T{graf}{.18} \T{spraying}{.18} \T{tagging}{.16} \T{bombing}{.15} & Active vandalism involving spray painting actions \\ \addlinespace
5  & \T{graffiti}{.23} \T{streetart}{.18} \T{banksy}{.18} \T{mural}{.17} & Large-scale murals with Banksy-style aesthetics \\ \addlinespace
6  & \T{graff}{.16} \T{colourful}{.13} \T{scribed}{.13} \T{tagging}{.13} & Chaotic wall markings with colorful scribbles \\ \addlinespace
11 & \T{urbanart}{.20} \T{sprayed}{.18} \T{graffiti}{.17} \T{grasso}{.17} & Urban decay imagery featuring painted walls \\ \addlinespace
15 & \T{graffiti}{.18} \T{paint}{.14} \T{\twemoji{artist palette}}{.14} \T{wallart}{.14} & Artistic materials focusing on paint and palettes \\ \midrule

% --- Group 4 ---
\Grp{Group IV: Criminal Identity}
3 & \T{sith}{.16} \T{cloak}{.14} \T{anonymous}{.12} \T{taliban}{.12} & Threatening figures with obscured faces or hoods \\ \addlinespace
4 & \T{incense}{.14} \T{anonymous}{.14} \T{masks}{.13} \T{liquor}{.12} & Masked individuals in looting or riot contexts \\ \addlinespace
1 & \T{shifter}{.12} \T{jewelry}{.12} \T{necklace}{.12} \T{knuckle}{.10} & Stolen valuables and improvised weapons \\
\bottomrule
\end{tabularx}
\end{table*}

\begin{table*}[h]
\centering
\caption{Semantic Analysis of Prototypes Learned for Concept ``Harassment''}
\label{tab:harassment_prototypes_grouped}
\renewcommand{\arraystretch}{1.1}

\newcommand{\T}[2]{\textbf{#1}~\scriptsize{(#2)}}
\newcommand{\Grp}[1]{\multicolumn{3}{c}{\textit{\textbf{#1}}} \\ \cmidrule(lr){1-3}}

\begin{tabularx}{\textwidth}{@{}c >{\raggedright\arraybackslash}X >{\raggedright\arraybackslash}X@{}}
\toprule
\textbf{ID} & \textbf{Top Associated Tokens} & \textbf{Semantic nuances} \\ \midrule

% --- Group 1: Aggression ---
\Grp{Group I: Overt Aggression \& Emotional Outbursts}
1 & \T{hysteria}{.17} \T{scream}{.16} \T{yell}{.16} \T{rage}{.15} & Intense emotional volatility and verbal abuse \\ \addlinespace
12 & \T{anger}{.15} \T{avoid}{.14} \T{brawl}{.13} \T{bullies}{.12} & Physical confrontation and heated conflicts \\ \addlinespace
14 & \T{enthusi}{.15} \T{roar}{.15} \T{yell}{.15} \T{shout}{.14} & Loud vocal aggression and intimidation \\ \addlinespace
5 & \T{scream}{.15} \T{dubstep}{.15} \T{hysteria}{.14} \T{phobia}{.14} & Auditory distress signals and chaotic noise \\ \midrule

% --- Group 2: Workplace/Institutional ---
\Grp{Group II: Workplace \& Professional Contexts}
2 & \T{boardroom}{.13} \T{conducting}{.12} \T{offices}{.12} \T{interviews}{.11} & Power dynamics in formal meeting environments \\ \addlinespace
13 & \T{deadlines}{.11} \T{adverts}{.11} \T{usability}{.11} \T{rage}{.11} & Stress-induced hostility in work or digital settings \\ \addlinespace
0 & \T{pancra}{.17} \T{monoxide}{.17} \T{chauffe}{.15} \T{attendant}{.14} & Service interactions involving drivers or staff \\ \midrule

% --- Group 3: Intrusion/Exclusion ---
\Grp{Group III: Intrusion, Exclusion \& Stalking}
4 & \T{photobomb}{.12} \T{paparazzi}{.12} \T{shy}{.12} \T{bullied}{.11} & Unwanted documentation and invasive attention \\ \addlinespace
7 & \T{standoff}{.15} \T{exclusion}{.13} \T{inspection}{.13} \T{glare}{.12} & Social exclusion and intimidating surveillance \\ \addlinespace
11 & \T{fleeing}{.13} \T{stride}{.13} \T{refused}{.12} \T{running}{.12} & Pursuit and evasion scenarios \\ \addlinespace
3 & \T{solitary}{.16} \T{commuter}{.16} \T{alone}{.15} \T{loneliness}{.15} & Targeting isolated individuals in public spaces \\ \midrule

% --- Group 4: Mockery/Ambiguous ---
\Grp{Group IV: Mockery \& Ambiguous Social Cues}
8 & \T{laughter}{.15} \T{laugh}{.14} \T{playful}{.12} \T{comedians}{.12} & Derisive laughter or humiliation disguised as humor \\ \addlinespace
15 & \T{smiles}{.15} \T{smiley}{.13} \T{joyful}{.13} \T{laughter}{.12} & Discomforting displays of forced friendliness \\ \addlinespace
10 & \T{joyous}{.13} \T{feliz}{.13} \T{ecstatic}{.12} \T{\twemoji{grin}}{.12}& Excessive expressiveness typical of online trolling \\ \addlinespace
6 & \T{pulsion}{.13} \T{angst}{.12} \T{refusal}{.11} \T{contempt}{.11} & Expressions of contempt and moody rejection \\
\bottomrule
\end{tabularx}
\end{table*}

\begin{table*}[h]
\centering
\caption{Semantic Analysis of Prototypes Learned for Concept ``Hate''}
\label{tab:hate_prototypes_grouped}
\renewcommand{\arraystretch}{1.1}

\newcommand{\T}[2]{\textbf{#1}~\scriptsize{(#2)}}
\newcommand{\Grp}[1]{\multicolumn{3}{c}{\textit{\textbf{#1}}} \\ \cmidrule(lr){1-3}}

\begin{tabularx}{\textwidth}{@{}c >{\raggedright\arraybackslash}X >{\raggedright\arraybackslash}X@{}}
\toprule
\textbf{ID} & \textbf{Top Associated Tokens} & \textbf{Semantic nuances} \\ \midrule

% --- Group 1: Explicit Hate ---
\Grp{Group I: Explicit Hate Speech \& Bias}
5 & \T{hated}{.14} \T{hates}{.14} \T{hate}{.13} \T{homophobia}{.11} & Direct verbal expressions of bias and homophobia \\ \addlinespace
3 & \T{hate}{.13} \T{iel}{.13} \T{hates}{.13} \T{hater}{.12} & General hate speech terms and text fragments \\ \addlinespace
15 & \T{fansign}{.15} \T{schild}{.13} \T{hated}{.13} \T{lettering}{.12} & Hate slogans displayed on signs or banners \\ \addlinespace
9 & \T{atar}{.16} \T{halt}{.16} \T{hayat}{.15} \T{hated}{.15} & Morphological variations of hate-related terms \\ \midrule

% --- Group 2: Nationalism ---
\Grp{Group II: Nationalist Imagery \& Extremism}
14 & \T{usair}{.15} \T{americafirst}{.15} \T{merica}{.15} \T{patriots}{.15} & Slogans related to 'America First' movements \\ \addlinespace
6 & \T{usair}{.16} \T{ussoccer}{.14} \T{usmnt}{.13} \T{patriotic}{.13} & Patriotism co-opted for exclusionary narratives \\ \addlinespace
2 & \T{usair}{.16} \T{merica}{.16} \T{ussoccer}{.15} \T{patriotic}{.15} & Heavy usage of national flags and identity markers \\ \midrule

% --- Group 3: Violence ---
\Grp{Group III: Civil Unrest \& Destruction}
10 & \T{riot}{.16} \T{riots}{.16} \T{extingui}{.15} \T{burn}{.15} & Scenes of civil disorder, protests, and arson \\ \addlinespace
0 & \T{molten}{.15} \T{\twemoji{fire}\twemoji{fire}\twemoji{fire}}{.14} \T{propane}{.13} \T{inciner}{.12} & Imagery of fire, burning, and incendiary devices \\ \midrule

% --- Group 4: Vandalism ---
\Grp{Group IV: Vandalism \& Hate Symbols}
13 & \T{graffiti}{.16} \T{rune}{.15} \T{monogram}{.15} \T{stencil}{.15} & Hate symbols (runes) sprayed on public walls \\ \addlinespace
7 & \T{graffiti}{.17} \T{stencil}{.17} \T{painter}{.17} \T{urbanart}{.15} & Defacement of property using stencils and paint \\ \addlinespace
1 & \T{tagging}{.15} \T{painter}{.15} \T{spraying}{.13} \T{graf}{.13} & Acts of tagging and street-level vandalism \\
\bottomrule
\end{tabularx}
\end{table*}

\begin{table*}[h]
\centering
\caption{Semantic Analysis of Prototypes Learned for Concept ``Shocking''}
\label{tab:shocking_prototypes_split}
\renewcommand{\arraystretch}{1.1}

\newcommand{\T}[2]{\textbf{#1}~\scriptsize{(#2)}}
\newcommand{\Grp}[1]{\multicolumn{3}{c}{\textit{\textbf{#1}}} \\ \cmidrule(lr){1-3}}

\begin{tabularx}{\textwidth}{@{}c >{\raggedright\arraybackslash}X >{\raggedright\arraybackslash}X@{}}
\toprule
\textbf{ID} & \textbf{Top Associated Tokens} & \textbf{Semantic nuances} \\ \midrule

% --- Group 1 (New): Realistic Gore ---
\Grp{Group I: Realistic Horror \& Physical Gore}
11 & \T{bloods}{.13} \T{butchers}{.13} \T{gruesome}{.12} \T{slaughter}{.12} & Graphic scenes of butchery and carnage \\ \addlinespace
8 & \T{bloody}{.12} \T{blooded}{.11} \T{zombie}{.11} \T{gore}{.11} & Visceral imagery focusing on blood and injury \\ \midrule

% --- Group 2 (New): Fictional Horror ---
\Grp{Group II:  Fictional Horror}
0 & \T{zombies}{.09} \T{walkingdead}{.09} \T{carnage}{.09} \T{butchers}{.08} & Horror fiction tropes, specifically decaying zombies \\ \addlinespace
5 & \T{hellboy}{.08} \T{pollock}{.08} \T{zombies}{.07} \T{blood}{.06} & Stylized violence, comic art, or splatter effects \\ \midrule

% --- Group 3: Ideological ---
\Grp{Group III: Political Controversy \& Polarization}
2 & \T{kaepernick}{.09} \T{supremacist}{.09} \T{gopdebate}{.08} \T{zionist}{.08} & Highly controversial political figures and protests \\ \addlinespace
14 & \T{nollywood}{.09} \T{nigerians}{.09} \T{blackhistorymonth}{.08} \T{farrakhan}{.06} & Activism and racially charged imagery \\ \midrule

% --- Group 4: Spectacle (Merged) ---
\Grp{Group IV: High-Risk Stunts, Spectacle \& Exposure}
13 & \T{daredevil}{.11} \T{torque}{.11} \T{varan}{.11} \T{sledge}{.10} & Adrenaline-inducing stunts and dangerous feats \\ \addlinespace
12 & \T{oahu}{.13} \T{seychel}{.13} \T{saltlife}{.13} \T{sunbathing}{.12} & Revealing beachwear and body exposure \\ \addlinespace
9 & \T{bedside}{.13} \T{bedroom}{.12} \T{bed}{.10} \T{boudo}{.10} & Intimacy and invasion of private spaces \\ \addlinespace
3 & \T{caricature}{.09} \T{lynda}{.09} \T{lum}{.09} \T{pinup}{.08} & Exaggerated caricatures and bizarre visual art \\ \addlinespace
15 & \T{kaitlyn}{.16} \T{jiu}{.15} \T{nightlife}{.12} \T{nightw}{.11} & Flashy nightlife scenes and celebrity culture \\
\bottomrule
\end{tabularx}
\end{table*}

\begin{table*}[h]
\centering
\caption{Semantic Analysis of Prototypes Learned for Concept ``Self-Harm''}
\label{tab:selfharm_prototypes_grouped}
\renewcommand{\arraystretch}{1.1}

\newcommand{\T}[2]{\textbf{#1}~\scriptsize{(#2)}}
\newcommand{\Grp}[1]{\multicolumn{3}{c}{\textit{\textbf{#1}}} \\ \cmidrule(lr){1-3}}

\begin{tabularx}{\textwidth}{@{}c >{\raggedright\arraybackslash}X >{\raggedright\arraybackslash}X@{}}
\toprule
\textbf{ID} & \textbf{Top Associated Tokens} & \textbf{Semantic nuances} \\ \midrule

% --- Group 1: Medication ---
\Grp{Group I: Medication, Overdose \& Substance Use}
5 & \T{capsule}{.18} \T{capsules}{.17} \T{supplements}{.15} \T{intment}{.15} & Pills and pharmaceutical imagery suggesting overdose \\ \addlinespace
2 & \T{octane}{.17} \T{capsule}{.17} \T{sesh}{.17} \T{antidote}{.16} & Drug paraphernalia and antidotes \\ \addlinespace
4 & \T{gazette}{.16} \T{capsu}{.16} \T{sesh}{.16} \T{drug}{.15} & References to drugs and doping substances \\ \midrule

% --- Group 2: Depression/Sorrow ---
\Grp{Group II: Melancholy, Isolation \& Emotional Distress}
6 & \T{dency}{.19} \T{burroughs}{.19} \T{refused}{.19} \T{melancholy}{.19} & Themes of sorrow, insomnia, and bruising \\ \addlinespace
13 & \T{refused}{.18} \T{syrian}{.18} \T{cist}{.18} \T{sore}{.16} & Physical pain (sore, prick) and rejection \\ \addlinespace
1 & \T{notte}{.20} \T{refused}{.19} \T{noches}{.19} \T{wexmondays}{.18} & Dark, nocturnal imagery (notte/noches) and isolation \\ \addlinespace
9 & \T{endon}{.19} \T{wexmondays}{.19} \T{notte}{.18} \T{refused}{.18} & Depressive aesthetics and urban alienation \\ \midrule

% --- Group 3: Institutional ---
\Grp{Group III: Institutional Settings \& Intervention}
11 & \T{restroom}{.20} \T{elevator}{.20} \T{olulu}{.18} \T{corridor}{.18} & Sterlie, confined spaces like hospital hallways \\ \addlinespace
12 & \T{polici}{.15} \T{policeman}{.14} \T{policemen}{.14} \T{officer}{.14} & Emergency responders and police intervention \\ \addlinespace
0 & \T{drummond}{.12} \T{constable}{.12} \T{confit}{.11} \T{condemned}{.10} & Authority figures and feelings of condemnation \\ \midrule

% --- Group 4: Labels/Symbols ---
\Grp{Group IV: Warning Labels \& Medical Symbols}
7 & \T{sticker}{.17} \T{schild}{.17} \T{indication}{.16} \T{label}{.15} & Warning stickers or medical labels \\ \addlinespace
10 & \T{label}{.15} \T{sticker}{.14} \T{logo}{.14} \T{schild}{.14} & Badges and emblems associated with institutions \\ \addlinespace
14 & \T{iu}{.16} \T{saber}{.15} \T{octane}{.15} \T{hazmat}{.14} & Hazardous material warnings and danger signs \\
\bottomrule
\end{tabularx}
\end{table*}

\begin{table*}[h]
\centering
\caption{Semantic Analysis of Prototypes Learned for Concept ``Violence''}
\label{tab:violence_prototypes_grouped}
\renewcommand{\arraystretch}{1.1}

\newcommand{\T}[2]{\textbf{#1}~\scriptsize{(#2)}}
\newcommand{\Grp}[1]{\multicolumn{3}{c}{\textit{\textbf{#1}}} \\ \cmidrule(lr){1-3}}

\begin{tabularx}{\textwidth}{@{}c >{\raggedright\arraybackslash}X >{\raggedright\arraybackslash}X@{}}
\toprule
\textbf{ID} & \textbf{Top Associated Tokens} & \textbf{Semantic nuances} \\ \midrule

% --- Group 1: Unrest ---
\Grp{Group I: Civil Unrest, Protest \& Siege}
0 & \T{protester}{.17} \T{riots}{.16} \T{woodstock}{.15} \T{beirut}{.14} & Large-scale riots, protests, and chaotic crowds \\ \addlinespace
10 & \T{woodstock}{.16} \T{protester}{.16} \T{shakur}{.16} \T{riots}{.14} & Counterculture movements and civil disobedience \\ \addlinespace
9 & \T{waco}{.14} \T{civilian}{.13} \T{escobar}{.13} \T{sarajevo}{.13} & Historical sieges and conflict zones \\ \midrule

% --- Group 2: Crime/Police ---
\Grp{Group II: Criminality \& Law Enforcement}
11 & \T{policeman}{.13} \T{civilian}{.12} \T{waco}{.12} \T{dystopian}{.12} & Police presence and state enforcement scenes \\ \addlinespace
14 & \T{banksy}{.15} \T{escobar}{.14} \T{robber}{.14} \T{criminal}{.14} & Criminal archetypes (robbers) and outlaws \\ \addlinespace
4 & \T{goalscorer}{.14} \T{hardrock}{.14} \T{floyd}{.13} \T{escobar}{.13} & Aggressive male figures associated with crime or fighting \\ \midrule

% --- Group 3: Sports ---
\Grp{Group III: Contact Sports \& Physical Combat}
2 & \T{stadiums}{.13} \T{cheerleader}{.12} \T{kickoff}{.12} \T{rocky}{.11} & High-impact contact sports like rugby and football \\ \addlinespace
8 & \T{bloody}{.12} \T{slaughter}{.11} \T{carnage}{.11} \T{gore}{.11} & Graphic imagery of blood and bodily injury \\ \addlinespace
3 & \T{johnnie}{.11} \T{rockabilly}{.10} \T{rocky}{.10} \T{muaythai}{.09} & Martial arts and cinematic fighting references \\ \midrule

% --- Group 4: Noir/Subculture ---
\Grp{Group IV: Urban Noir, Nightlife \& Rebellion}
5 & \T{bartender}{.18} \T{waitress}{.15} \T{hopper}{.14} \T{bar}{.13} & Alcohol-fueled environments and bar scenes \\ \addlinespace
13 & \T{rockabilly}{.13} \T{morrissey}{.13} \T{streetcar}{.13} \T{scorsese}{.12} & Gritty aesthetic inspired by mob movies (Scorsese) \\ \addlinespace
7 & \T{sidewalks}{.15} \T{blackandwhitephotography}{.15} \T{refused}{.13} \T{protester}{.13} & Moody urban street photography indicating tension \\
\bottomrule
\end{tabularx}
\end{table*}

\end{document}